\documentclass[manuscript, screen]{acmart}
\usepackage{graphicx}
\usepackage{array}
\usepackage{amsmath}
\usepackage{float}
\usepackage{color}

\usepackage{booktabs}
\usepackage{multirow}

\usepackage{algorithm}
\usepackage{algorithmic}

\usepackage{amssymb}
\usepackage{booktabs}
\usepackage{caption}
\usepackage{bbding}
\usepackage{overpic}
\usepackage{subfigure}
\usepackage{makecell}
\usepackage{array}
\usepackage{colortbl,xcolor}

\setcopyright{acmcopyright}
\acmDOI{XXXXXXX.XXXXXXX}

\acmJournal{CSUR}

\newcommand{\etal}{\textit{et al}. }
\newcommand{\ie}{\textit{i.e., }}
\newcommand{\eg}{\textit{e.g., }}
\newcommand{\etc}{\textit{etc. }}

\definecolor{Gray}{gray}{0.8}

\DeclareRobustCommand\onedot{\futurelet\@let@token\@onedot}
\makeatother

\newcommand{\corrmark}{\textsuperscript{*}}
\newcommand{\corrnote}{\footnotetext[1]{\textsuperscript{*}Corresponding author.}}
\begin{document}

\title{Deep Face Restoration: A Survey}

\author{Tao Wang}
\email{taowangzj@gmail.com}
\affiliation{%
  \institution{The State Key Lab for Novel Software Technology,
Nanjing University}
  \city{Nanjing}
  \country{China}
}

\author{Kaihao Zhang\corrmark}
\affiliation{%
  \institution{Australian National University}
   \city{Canberra}
  \country{Australia}}
\email{super.khzhang@gmail.com}




\author{Jiankang Deng}
\affiliation{%
  \institution{Imperial College London}
  \city{London}
  \country{UK}}
\email{j.deng16@imperial.ac.uk}

\author{Tong Lu}
\email{lutong@nju.edu.cn}
\affiliation{%
  \institution{The State Key Lab for Novel Software Technology,
Nanjing University}
  \city{Nanjing}
  \country{China}
}


\author{Wei Liu}
\affiliation{%
  \institution{Tencent}
  \city{Shenzhen}
  \country{China}}
\email{wl2223@columbia.edu}


\author{Stefanos Zafeiriou}
\affiliation{%
  \institution{Imperial College London}
  \city{London}
  \country{UK}}
\email{s.zafeiriou@imperial.ac.uk}






\renewcommand{\shortauthors}{Wang et al.}

\begin{abstract}
Face Restoration (FR) aims to restore High-Quality (HQ) faces from Low-Quality (LQ) input images, which is a domain-specific image restoration problem in the low-level computer vision area. The early face restoration methods mainly use statistical priors and degradation models, which are difficult to meet the requirements of real-world applications in practice. In recent years, face restoration has witnessed great progress after stepping into the deep learning era. However, there are few works to systematically study the deep learning based face restoration methods. Thus, in this paper, we provide a comprehensive survey of recent advances in deep learning techniques for face restoration. Specifically, we first summarize different problem formulations and analyze the characteristics of face images. Second, we discuss the challenges of face restoration. With regard to these challenges, we present a comprehensive review of recent FR methods, including prior-based methods and deep-learning methods. Then, we explore developed techniques in the task of FR covering network architectures, loss functions, and benchmark datasets. We also conduct a systematic benchmark evaluation on representative methods. Finally, we discuss the future
directions including network designs, metrics, benchmark datasets, applications, \etc We also provide an open source repository for all the discussed methods, which is available at \url{https://github.com/TaoWangzj/Awesome-Face-Restoration}.
\end{abstract}

\ccsdesc[500]{Computing methodologies~Computational photography}


\keywords{Face restoration, Deep learning, Survey, Low-level vision, Facial prior}


\maketitle
\corrnote 
\section{Introduction}\label{sec:introduction}
\begin{figure*}[t]
\begin{center}
  \begin{overpic}[width=\textwidth]{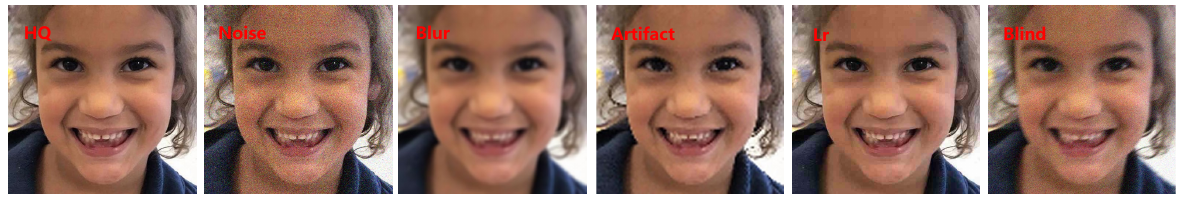} 
 \end{overpic}
 \vspace{-3mm}
 	\caption{Examples of low-quality face images degraded from a high-quality (HQ) image, with respect to noise, blur, artifact, low resolution, and a mix of the above factors.}
	\label{fig:examples}
	\end{center}
	\vspace{-7mm}
\end{figure*}

Face restoration, a domain-specific image restoration problem, is a classic task in the fields of image processing and computer vision. Face restoration is to restore the high-quality face image $I_{hq}$ from the degraded face image $I_{lq}=\mathcal{D}(I_{hq};n_{\delta})$, where $\mathcal{D}$ is the noise-irrelevant degradation function, and $n_{\delta}$ is the noise. According to different forms of the degradation function $\mathcal{D}$, the face restoration task can be divided into five main categories: (1) \textbf{face denoising}, which refers to removing the noise (\eg Gaussian noise) contained in the face image \cite{luo2015adaptive,anwar2017category}, (2) \textbf{face deblurring}, which is to recover a latent sharp face image from a blurry face image caused by various factors such as camera shake or object motion \cite{shen2020exploiting,yasarla2020deblurring}, (3) \textbf{face super-resolution} (also known as face hallucination \cite{zhang2022edface,zengy2024implicit}), which aims to enhance quality and resolution of low-resolution facial images \cite{baker2000hallucinating,zhou2015learning}, (4) \textbf{face artifact removal},  which refers to recovering high-quality face images from the given low-quality face images with artifacts caused by lossy compression in the process of image storage and transmission \cite{yang2020hifacegan,zhang2022blind}, (5) \textbf{blind face restoration}, which aims at restoring high-quality face images from the low-quality ones without the knowledge of degradation types or parameters \cite{yang2020hifacegan,wang2021towards}. Fig.~\ref{fig:examples} illustrates exemplar low-quality face images caused by these forms of degradation, which influence not only the visual quality, but also the performance of down-stream computer vision algorithms.
Thus, face restoration has a wide range of applications, including face recognition~\cite{li2004floatboost}, privacy protection~\cite{yu2016iprivacy}, and autonomous driving~\cite{chen2015deepdriving}.

Early face restoration methods mainly focus on statistic prior and degradation models, which can be approximately divided into Bayesian inference based methods~\cite{baker2000hallucinating,tang2003face}, subspace learning based methods~\cite{li2006illumination,he2003locality}, sparse representation based methods~\cite{wang2011face,tian2016weighted}, \etc In recent years, deep learning-based methods have attracted more and more attention with the development of deep learning and the availability of large-scale datasets. Thus, a large number of deep learning-based methods for face restoration have been proposed in the literature. Generally speaking, deep learning-based face restoration methods adopt different techniques to build state-of-the-art networks. The employed techniques mainly focus on the following aspects: different deep learning architectures~\cite{zhou2015learning,yu2016ultra,wang2022restoreformer,yu2017hallucinating,wang2023dr2}, different facial priors~\cite{chen2018fsrnet,li2018learning,gu2020image,yu2018face}, different loss functions~\cite{johnson2016perceptual,chen2018fsrnet,li2020enhanced,wang2017deepdeblur}, different learning strategies~\cite{menon2020pulse,li2020blind}, \etc Although deep learning solutions have dominated the research of face restoration in recent years, there is still a lack of in-depth and comprehensive surveys on face restoration with deep learning technology. Thus, this paper provides a comprehensive and systematic review of deep learning methods for the face restoration task.

\begin{figure*}[t]
\begin{center}
  \begin{overpic}[width=0.95\textwidth]{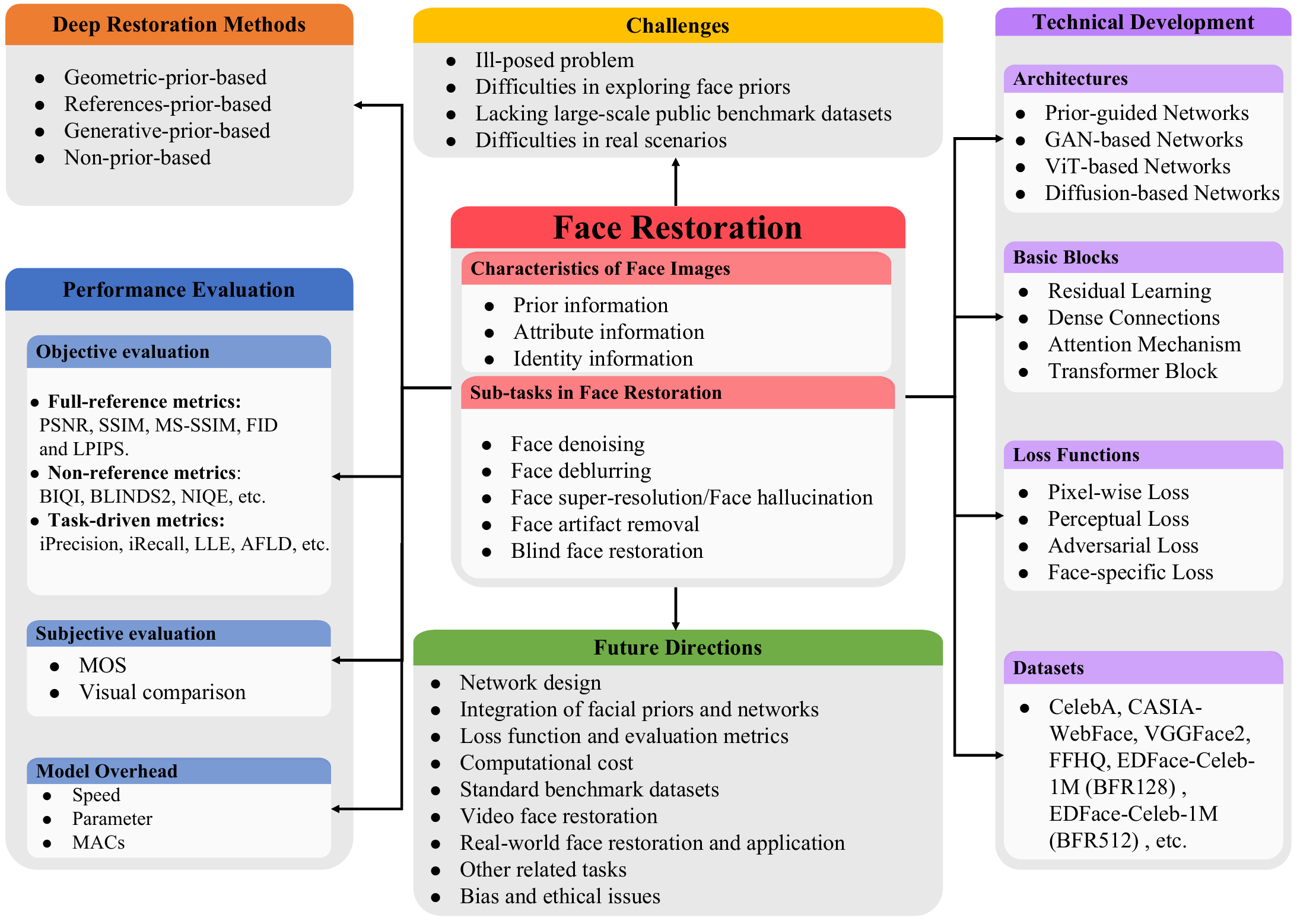} 
 \end{overpic}
 \vspace{-3mm}
	\caption{Taxonomy of this survey on face restoration using deep learning techniques.} 
	\label{fig:Taxonomy}
	\end{center}
	\vspace{-7mm}
\end{figure*}
\textbf{Differences from Other Related Reviews}. So far, there are few surveys about the overview of the face restoration task, though some surveys are related to the topic of face restoration. We divide them into three groups and discuss their differences in the following. (1) The first group~\cite{yang2020single,tian2020deep,wang2020deep,zhang2022deep,su2022survey,he2025diffusion} aims to discuss general image restoration using deep learning techniques. For example, in~\cite{yang2020single,tian2020deep,wang2020deep,zhang2022deep,xiang2023deep}, they discuss the common causes of one specific task in image restoration, such as deraining, denoising, super-resolution, and deblurring, respectively, and review different deep learning-based methods. \cite{su2022survey} pays more attention to reviewing deep learning methods for general image restoration tasks that include image deblurring, denoising, dehazing, and super-resolution. More recently, \cite{he2025diffusion} focuses on discussing the diffusion model used for general image restoration tasks. (2) The second group~\cite{liang2012survey,wang2014comprehensive,nguyen2018super,rajput2018face} focuses on reviewing the advances and development in traditional face super-resolution methods such as subspace learning based methods~\cite{li2006illumination,he2003locality}, and sparse representation based methods~\cite{wang2011face,tian2016weighted}. (3) The third group~\cite{liu2019survey,jiang2021deep} reviews the recent development in face super-resolution with deep learning techniques. Although the topic is related to ours, they focus only on the specific task of face super-resolution, whose scope is narrower than ours. In addition, a recent survey on deep face restoration~\cite{li2025survey} reviews deep face restoration methods from non-blind to blind settings. Compared with it, our survey further provides a more fine-grained and comprehensive coverage of face restoration sub-tasks, including face denoising, face deblurring, face super-resolution, face artifact removal, and blind face restoration, and summarizes recent advances with an emphasis on face-specific priors and practical challenges. Differently, our work systematically and comprehensively reviews recent advances in deep learning-based methods for face denoising, face deblurring, face super-resolution, face artifact removal, and blind face restoration.


\textbf{Our Contributions}. This work systematically and comprehensively reviews the research progress of face restoration technology in recent years. The taxonomy of this survey is shown in Fig.~\ref{fig:Taxonomy}. We conduct this survey in different aspects, including problem formulation, existing challenges, state-of-the-art methods, technical development, performance evaluation, and future directions. The contributions of this paper are summarized as follows. 
(I) We discuss the main degradation models in face restoration, the commonly used metrics, and the characteristics of face images that differ from natural images. (II) We discuss existing challenges in face restoration and provide a comprehensive overview of existing deep learning-based face restoration methods. (III) We provide in-depth analysis and discussion
about the technical development of the methods, covering network architectures, basic blocks, loss functions, and benchmark datasets. (IV) We conduct a benchmark study of representative methods on popular face benchmarks, which will facilitate future experimental comparisons. (V) We analyze the open challenges of the face restoration task and discuss its future directions to guide future research for the community.



\textbf{Review Methodology and Organization}. To make the review process transparent, we conducted a structured literature search covering the past decade (2015--2025) on Google Scholar, IEEE Xplore, ACM Digital Library, and arXiv using face-restoration-related keywords (\eg face super-resolution, face deblurring, face denoising, face artifact removal, blind face restoration, GAN/diffusion, and face priors). We selected studies via title/abstract and full-text screening, excluded non-face or insufficient-detail works, and complemented the collection via citation snowballing from representative papers. The remainder of this paper is organized as follows. In Section~\ref{sec:background}, we successively introduce the problem definitions of six common face restoration tasks, the image quality evaluation metrics, and the characteristics of face images. In Section~\ref{sec:literature_Survey}, we discuss the challenges of face restoration and analyze how existing face restoration methods address these challenges. Section~\ref{sec:development_review} reviews the technical development of deep face restoration, including network architectures, basic blocks, loss functions, and datasets. Section~\ref{sec:performance_evaluation} reports the experimental results of existing methods. In Section~\ref{sec:future_directions}, we discuss the future directions of face restoration. Finally, Section~\ref{sec:conclusion} concludes this paper.

\section{Background} \label{sec:background}
This section introduces the background of deep face restoration. We first summarize common tasks and degradations, then review image quality assessment metrics, and finally discuss face-specific characteristics and attribute information that motivate face priors and influence face restoration and evaluation.

\textbf{Problem Formulation}. Image degradation during formation, transmission, and storage can take various forms in real-world facial images, including additive noise, spatially invariant or variant blur, aliasing, and compression artifacts. Generally, the degradation model is formulated as:
\begin{equation}
\label{eq:degradation_model}
I_{lq}=\mathcal{D}(I_{hq};n_{\delta}), 
\end{equation}
where $I_{lq}$ is the low-quality face image, $\mathcal{D}$ refers to the degradation function, $I_{hq}$ is the corresponding high-quality face image, and $n_{\delta}$ usually denotes additive white Gaussian noise with a noise level $\delta$.
By specifying different $\mathcal{D}$, one can get different degradation. For example, noise degradation~\cite{anwar2019real,yue2019variational} that $\mathcal{D}$ is an identity function. Blur degradation~\cite{kupyn2018deblurgan,zhang2022deep} where $\mathcal{D}$ is a convolution/averaging operation. Low-resolution degradation~\cite{yang2019deep,dong2015image,lim2017enhanced,wang2018esrgan} when $\mathcal{D}$ is a combination of the convolution and downsampling operations. Artifact degradation~\cite{dong2015compression,liu2020comprehensive} when $\mathcal{D}$ is a JPEG compression operation. Mixed degradation~\cite{yang2020hifacegan,wang2021towards} when $\mathcal{D}$ is a combo of various factors.

FR refers to the recovery of a high-quality face image from its degraded low-quality counterpart. Namely, it aims to find the inverse of the degradation model in Eq. \ref{eq:degradation_model} as:
\begin{equation}
I_{hq}=\mathcal{D}^{-1}(I_{lq};n_{\delta}), 
\end{equation}
where $\mathcal{D}^{-1}$ is the face restoration model. If the degradation factors are provided, the FR task is regarded as non-blind face restoration, like face denoising, face deblurring, face super-resolution, and face artifact removal. Otherwise, the FR task is called blind face restoration. In the following, we detail the specific problem definition of sub-tasks in FR, where we mainly introduce some commonly used degradation models.

\textit{Face Denoising}. This sub-task focuses on removing noise from an observed noisy face image. The noisy face image is typically constructed by the additive model, which is formulated as:
\begin{equation}
I_{n}=I_{c}+n_{\delta},
\end{equation}
where $I_{c}$, $I_{n}$, and $n_{\delta}$ represent the clean face image, noisy face image, and additive Gaussian noise with a noise level $\delta$, respectively. Face Denoising is to find the inverse of the degradation model.

\textit{Face Deblurring}. Face blur is a common problem in captured face images. It mainly contains motion blur~\cite{zhang2020deblurring} caused by the relative movement between the object and the camera, out-of-focus blur~\cite{chen2015multispectral} caused by the misalignment between the target and the camera focus. Face deblurring mainly considers motion blur, which can be modeled as:
\begin{equation}
I_{b}=k_{\sigma}*I_{s}+n_{\delta},
\end{equation}
where $I_{b}$ is the blurry face image, $I_{s}$ is the sharp face image, $k_{\sigma}$ is the blur kernel, $*$ is the convolution operation, and $n_{\delta}$ is the additive noise. Face deblurring is to obtain the inverse function of the degradation model, so as to generate sharp face images.

\textit{Face Super-resolution}. As a domain-specific image super-resolution problem, face super-resolution refers to enhancing the resolution of low-resolution (LR) face images and producing high-resolution (HR) face images with rich details. The degradation model is formulated as:
\begin{equation}
I_{lr}= (I_{hr} * k_{\sigma}) \downarrow_{s}+n_{\delta},
\end{equation}
where $I_{lr}$ is the low-resolution face image, $I_{hr}$ is the high-resolution face image, $k_{\sigma}$ is the blur kernel, $*$ is the convolutional operation, $n_{\delta}$ is the noise, and $ \downarrow_{s}$ is the downsampling operation with a scale factor $s$. $s$ is usually set as 2, 3, 4, and 8 in the face super-resolution task. Based on the degradation, face super-resolution aims to simulate the inverse process of the degradation model and recover the HR face image from the LR face image.

\textit{Face Artifact Removal}. In real-world applications, lossy compression techniques (\eg JPEG, Webp, and HEVC-MSP) are widely adopted for saving storage space and bandwidth. However, lossy compression easily leads to information loss and introduces undesired artifacts for recorded face images. Given a high-quality face image $I_{hq}$, its compression process is as follows:
\begin{equation}
I_{lq}= J(I_{hq})+n_{\delta},
\end{equation}
where $I_{lq}$ is the compressed face image, $J$ denotes the image compression. As JPEG is the most extensively used way for image compression, researchers thus focus more on this type of degradation in the task of face artifact removal. According to the image compression process, face artifact removal is devoted to learning the inverse process of the degradation model and generating HQ face images.

\textit{Blind Face Restoration}. Unlike focusing on a single type of degradation, blind face restoration aims to handle severely degraded face images in the wild. The degradation of face images is complex in this task, which is a random combination of noise, blur, low resolution, and JPEG compression artifacts. The degradation model of blind face restoration can be defined as:
\begin{equation}
I_{lq}= \left\{J P E G_{q}((I_{hq} * k_{\sigma}) \downarrow_{s}+n_{\delta})\right\}\uparrow_{s},
\label{degradation_model}
\end{equation}
where $*$ is the convolution operation, $k_{\sigma}$ is the blur kernel, $J P E G_{q}$ is JPEG compression function with quality factor $q$, $\downarrow_{s}$ is downsampling operation with scaling factor $s$, $n_{\delta}$ is the noise, and $\uparrow_{s}$ is upsampling operation with scaling factor $s$. The goal of blind face restoration is to recover HQ face images by modeling the inverse process of the above degradation model.

\textbf{Image Quality Assessment}. Evaluating the quality of restored images is essential. Image quality assessment methods are generally categorized into subjective and objective approaches. Subjective evaluation, such as the Mean Opinion Score (MOS)~\cite{hossfeld2016qoe}, involves human raters assigning visual scores. While accurate, this method is costly and time-consuming. Objective evaluation is more practical and can be divided into full-reference, no-reference, and task-driven metrics. \textit{Full-reference metrics} compare the restored image with its ground truth. Common metrics include PSNR~\cite{hore2010image}, SSIM~\cite{wang2004image}, MS-SSIM~\cite{wang2003multiscale}, and LPIPS~\cite{zhang2018unreasonable}. PSNR measures pixel-wise differences, while SSIM also considers luminance, contrast, and structure. MS-SSIM enhances SSIM by aggregating local similarities. Unlike these pixel-based metrics that often favor overly smooth results, LPIPS evaluates perceptual similarity aligned with human vision. \textit{No-reference metrics} estimate quality without ground truth. Widely used ones in face restoration (FR) include BIQI~\cite{moorthy2010two}, BLINDS2~\cite{saad2012blind}, BRISQUE~\cite{mittal2012no}, CORNIA~\cite{ye2012unsupervised}, DIIVINE~\cite{moorthy2011blind}, SSEQ~\cite{liu2014no}, NIQE~\cite{mittal2012making}, and FID~\cite{heusel2017gans}. Among them, NIQE and FID~\cite{wang2021towards,wang2022restoreformer,zhang2022blind} are commonly adopted to assess the naturalness of restored faces. \textit{Task-driven metrics}, specific to face restoration, consider identity-related features. Examples include iPrecision, iRecall~\cite{zhao2022rethinking}, LLE~\cite{yang2020hifacegan}, Deg~\cite{wang2021towards}, AFLD, and AFICS~\cite{zhang2022blind}, which evaluate fidelity using landmarks, face IDs, or identity similarity.

\textbf{Analysis of Face Image}. As we can see, the captured face images contain a wide variety of information related to humans, such as human face geometry spatial distribution information. Thus, different from the general image restoration task, 
the face geometry information (\ie facial prior) can be exploited for face restoration. In the past decades, a large amount of human face information in face images has been explored to assist in face restoration. Generally speaking, the information in the face image can be divided into three categories: human attribute information, human identity information, and other prior information. We introduce them as follows.
\begin{figure*}[t]
\begin{center}
  \begin{overpic}[width=\textwidth]{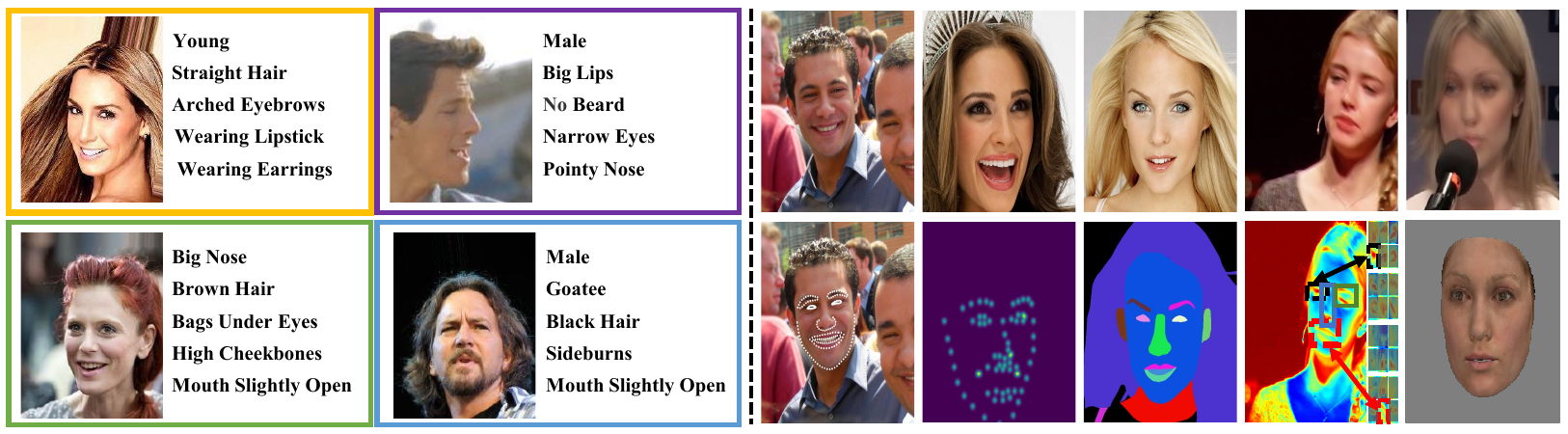} 
   \end{overpic}
    \vspace{-3mm}
    \caption{Left: Illustration of typical attributes in face images. Right: Examples of common face priors. The top row shows face images, while the bottom row presents the corresponding priors, including facial landmarks~\cite{chen2018fsrnet,kim2019progressive}, facial heatmaps~\cite{yu2018face}, facial parsing maps~\cite{chen2021progressive}, facial dictionaries~\cite{li2020blind}, and 3D face priors~\cite{hu2021face}.}
	\label{fig:attribute}
	\end{center}
	\vspace{-6mm}
\end{figure*}

   

\textit{Human Attribute Information}. As illustrated in Fig. \ref{fig:attribute}, a face image usually contains special attributes of a figure, such as gender, age, glasses, emotion \etc These face affiliated attributes are beneficial to multiple tasks, including face recognition~\cite{tu2021joint,deng2020retinaface}, face verification~\cite{kumar2009attribute}, and face restoration~\cite{liu2021cg,xia2021towards}. As the degradation in face images is rather diverse and complex, it is difficult for the restoration model to recover the clear image relying only on the degraded image. Therefore, some methods~\cite{dogan2019exemplar,wang2021towards,liu2021cg} exploit human attributes in the face image as additional information to guide the restoration process. For example, Liu \etal \cite{liu2021cg} introduce the class-attribute information into the local detail restoration stage to further enhance local details.

\textit{Human Identity Information}. In addition to the basic attributes in the face image, each face has its unique identity information. The identity information can be used to guide the model to generate faces close to the real identity. On the one hand, humans' accurate perception of the face mainly depends on the identity information of the face. On the other hand, adopting only the pixel-level loss to supervise the restoration model training cannot produce accurate identity-related facial details for the task of face restoration. For example, \cite{menon2020pulse} can generate face images with high perceptual quality. However, it can not retain the face identity well in the recovered images. Thus, the identity information is introduced to improve both the recognizability and performance of face restoration in the literature~\cite{zhang2018super,grm2019face,chen2020identity}.

\textit{Other Prior Information}. As illustrated in Fig. \ref{fig:attribute}, some representative facial priors in face restoration are facial landmarks~\cite{chen2018fsrnet,kim2019progressive}, facial heatmaps~\cite{yu2018face}, facial parsing maps~\cite{chen2021progressive}, and 3D face prior~\cite{hu2021face}. 1) Facial landmarks. There are some important reference points of facial components, such as eye centers, nose tips, and mouth corners of humans in the image. Different datasets provide different numbers of facial landmarks for each face image. For instance, CelebA dataset contains $5$ landmarks~\cite{liu2015deep}, FFHQ dataset includes $68$ landmarks~\cite{karras2019style}, and Helen dataset provides $194$ landmarks~\cite{le2012interactive}. In addition, various facial landmark detection methods~\cite{dong2018style,wu2017facial} can help detect landmarks.
2) Facial heatmaps. Compared to facial landmarks directly providing reference points of facial components, facial heatmaps describe the probability that reference points are facial landmarks. Specifically, based on the facial landmarks,  each landmark is encoded by using a 2D Gaussian centered at the coordinates of that landmark to generate the facial heatmaps. 3) Facial parsing maps. These are semantic feature maps of face images, which are separated out face components (\eg nose, skin, eyes, and hair) from the face images. 4) 3D face prior. In contrast to the 2D prior without considering high-dimensional information (\eg position and shape of faces), the 3D face prior is developed for face restoration~\cite{hu2021face}. 3D face prior provides rich 3D knowledge based on the fusion of different face attributes (\eg identity, facial expression, illumination, and face pose). In addition, reference-based prior~\cite{li2018learning,li2020blind} and generative prior~\cite{gu2020image,wang2021towards} are introduced in the literature to guide the face restoration models.

\section{Literature Survey} \label{sec:literature_Survey}
In this section, we first briefly analyze the challenges in face restoration tasks. Then, we present a systematic overview of face image restoration methods, including prior based deep restoration methods and non-prior based deep restoration methods.

 \subsection{Face Restoration Challenges}\label{sec:challenges} 
As a domain-specific image restoration task, face restoration aims to remove various unknown degradations in the low-quality face image and construct a high-quality one. However, there are several challenges in the task of face restoration.

\textbf{Ill-posed problem}. Although most existing methods are specifically developed for dealing with a single face restoration task, it is still an ill-posed problem, as the degradation types and degradation parameters of low-quality face images are unknown in advance. On the other hand, in practical scenarios, the degradation of face images is complex and diverse. Thus, designing effective and robust face restoration models to restore clear face images is a challenging problem.
  
\textbf{Difficulties in exploring face priors}. As a domain-specific image restoration task, some facial priors can be explored for face image restoration. Typical facial priors used in the literature include 1D vectors (identity and attributes), 2D images (facial landmarks, facial heatmaps, and facial parsing maps), and 3D prior. However, it is difficult to exploit the prior knowledge, because facial priors such as facial components and facial landmarks are usually extracted or estimated from low-quality images, which may be inaccurate and therefore directly affect the restoration performance. On the other hand, real-world low-quality images often contain complex and diverse degradation, and it is difficult to find appropriate priors to assist the process of face restoration. In addition, face restoration is different from image restoration due to the specialty of face images. 
For example, human eyes are more sensitive to face artifacts, bearing in mind a strong expectation of human face structure. Thus, it brings another difficulty in restoring the human face.

\textbf{Lacking large-scale public benchmark datasets}. With the development of deep learning techniques, deep learning-based methods have shown impressive performance in face restoration. Most deep learning-based face restoration methods strongly rely on large-scale datasets to train networks. However, most of the current face restoration methods are trained or tested on non-public datasets. Especially in their experiments, these methods usually synthesize low-quality images using their private schemes based on the high-quality images and randomly split them for training and evaluation respectively~\cite{zhou2015learning,cao2017attention,wang2017deepdeblur}. Though some works use fixed training/testing sets~\cite{li2020blind,gu2020image,wang2021towards,yang2020hifacegan}, the synthesized low-quality images are still different due to random noise, random combinations of degradation factors, \etc Therefore, it is still difficult to directly compare existing methods based on the reported results. Lacking public datasets directly leads to unfair performance comparison. In addition, lacking high-quality and large-scale benchmarks limits the potential of models. Therefore, it is a challenge to build more proper benchmark datasets for face restoration.

\begin{figure*}[t] 
\begin{center}
  \begin{overpic}[width=0.99\textwidth]{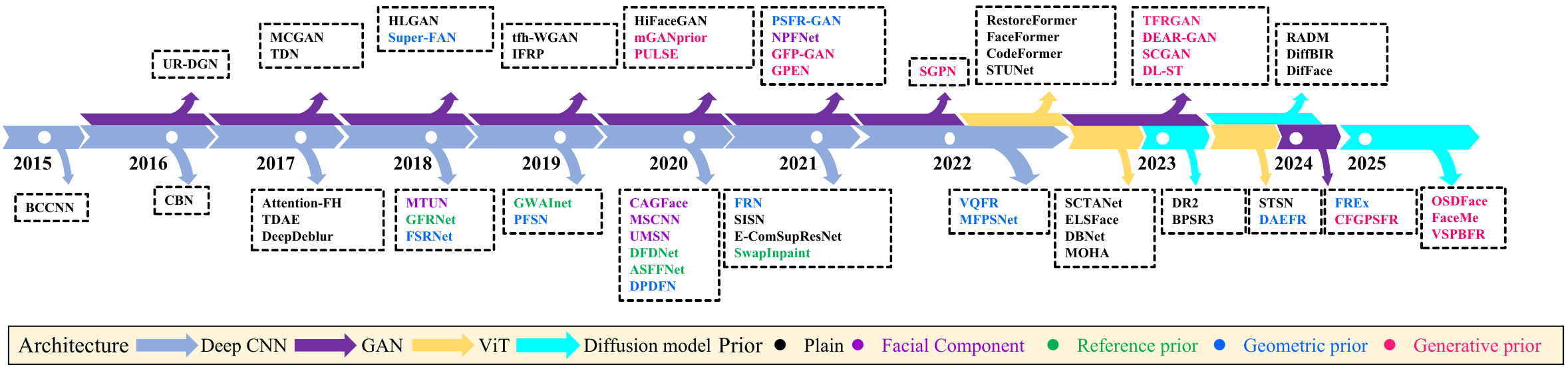}

  
   \end{overpic}
    \vspace{-3mm}
	\caption{Milestones of deep learning-based face restoration methods. We summarize the methods by different network architectures and facial priors. We list their names in the figure.}
	\label{fig:milestone}  
	\end{center}
	\vspace{-7mm}
\end{figure*}  

\textbf{Difficulties in real-world scenarios}. Though deep learning methods have acquired state-of-the-art performance in face restoration, most of them work in a supervised manner. Specifically, these approaches require a paired (low-quality and high-quality image pair) dataset, and they would fail if the conditions are unsatisfied. However, it is difficult to collect large-scale datasets with real paired samples in the real world due to the complex and changeable scene. Therefore, most methods synthesize low-quality images by degradation models to approximate the real low-quality image. In addition, the synthesized low-quality face images are probably less informative and inconsistent with real-world images. The models trained on synthetic data sets can easily lead to domain drift, which limits the applicability of the model in real scenarios.

In the following, we will introduce and analyze in detail how existing face restoration methods deal with the above challenges. 

\subsection{Face Restoration Methods}\label{sec:methods}

General image restoration methods aim to design efficient methods for recovering sharp natural images. However, as a highly structured object, the human face has specific characteristics that are ignored by general image restoration methods. Thus, most face restoration methods incorporate face prior knowledge to recover facial images with clearer facial structure. The developed face-specific priors in the models are mainly based on common sense that human faces exhibit small variations in a controlled environment. On the other hand, other methods aim to develop networks learning a mapping function between the low-quality and high-quality face images without the facial prior. The milestones of face restoration in the past years are illustrated in Fig.~\ref{fig:milestone}. We divide face restoration methods into two categories: prior based deep restoration methods and non-prior based deep learning approaches.
In addition, prior based deep restoration methods can be approximately divided into three sets: geometric prior based deep restoration methods, reference prior based deep restoration methods, and generative prior based deep restoration methods. In this part, we first summarize prior-based methods, including geometric prior, reference prior, and generative prior based designs. We then introduce non-prior based methods that learn restoration without explicit facial priors. For each group, we provide short summaries to highlight key ideas and limitations. In the following, we discuss these methods in detail. 

\textbf{Geometric Prior Based Deep Restoration Methods}.
This group mainly adopt the unique geometry and spatial distribution information of faces in the image to help the model progressively restore high-quality face images. Typical geometric priors include facial landmarks~\cite{chen2018fsrnet,kim2019progressive}, facial heatmaps~\cite{yu2018face} and facial parsing maps~\cite{chen2021progressive}. Chen \etal \cite{chen2018fsrnet} make the first attempt to design a specific face geometric prior estimation sub-network in a deep network and train them in an end-to-end manner for the face super-resolution task. Specifically, they first use a coarse network to recover the coarse high-resolution image. Then the coarse image is sent to a fine super-resolution network and a prior information estimation network to extract image features and estimate landmark heatmaps and parsing maps respectively. In the end, both image features and geometric prior are fed to a fine super-resolution decoder to restore high-resolution images. This pioneering work improves the performance of face super-resolution while also providing a solution to estimating geometric prior directly from low-quality face images. Another representative work is Super-FAN proposed by Bulat and Tzimiropoulos~\cite{bulat2018super}. Super-FAN is the first end-to-end system to simultaneously achieve facial super-resolution and facial landmark localization. The core insight in Super-FAN is using the joint training strategy to guide the network to learn more face geometric information, which is achieved by integrating a face-alignment sub-network via heatmap regression and loss optimization. To further utilize facial attributes around the landmark for restoring facial details,  Kim \etal \cite{kim2019progressive} propose a lightweight face alignment network to generate facial landmark heatmaps for the face super-resolution network by a progressive training method. For example, by including the parsing map, an additional face prior is provided, which improves the network's performance in repairing the faces. Recently, Chen \etal \cite{chen2021progressive} propose a multi-scale progressive model for face restoration, which recovers low-quality face images in a coarse-to-fine manner by semantic-aware style transformation. 

Compared with accurately estimating face landmarks or semantic maps directly from low-quality images, it is easy to localize facial components (not landmarks). Thus, another line of geometric prior based methods~\cite{yu2018face,kalarot2020component,shen2020exploiting,yasarla2020deblurring,kim2021progressive} is to take advantage of features of the facial component in human faces, \eg eyes and nose, as the prior information. For example, MTUN~\cite{yu2018face} is a representative two-branch method. The first branch performs face super-resolution. The second branch estimates facial component heatmaps. It uses four heatmaps to represent the eyes, nose, mouth, and chin. MTUN shows that using facial component information from low-quality face images can further improve face restoration performance. Recently, Yu \etal \cite{yu2022multi} make the first attempt to use multi-geometric priors to guide face restoration. Their network leverages semantic parsing maps, facial heatmaps, facial dictionaries, and pixel-level degraded image information. In contrast to the previously mentioned methods that focus on using 2D priors, Hu \etal \cite{hu2021face} make the first attempt to embed 3D priors into networks for general face recovery tasks. Compared with 2D priors, 3D priors can integrate parameter descriptions of face attributes (\eg identity, facial expression, texture, illumination, and facial pose) to provide 3D morphological knowledge and further improve the face restoration performance.

\textbf{Reference Prior Based Deep Restoration Methods}. This group exploits reference information, such as an exemplar image or retrieved facial components, to provide high-frequency details that may be missing in the degraded input. Previous works exploit facial prior purely relying on a single degraded image. It is worth noting that the degradation process is generally highly ill-posed, which fails to obtain an accurate facial prior. Thus, several methods aim to guide the face restoration process by using the facial structure or facial component dictionaries obtained from additional high-quality face images as reference prior~\cite{li2018learning,li2020blind,dogan2019exemplar,li2020enhanced}. Some reference prior based methods utilize the additional information provided by a high-resolution
guiding image with the same identity. Li \etal \cite{li2018learning} and Dogan \etal \cite{dogan2019exemplar} mainly employ a fixed frontal high-quality reference for each identity to
provide additional identity-aware information to help the process of face restoration. Specifically, Li \etal \cite{li2018learning} propose a guided face restoration network called GFRNet. It consists of a warping sub-network WarpNet and a reconstruction sub-network RecNet. WarpNet generates a flow field to warp the reference image and correct the face pose and expression. RecNet takes the low-quality image and the warped guidance as input to recover a high-quality face image. Since the ground-truth flow field is unavailable, they introduce a landmark loss to train WarpNet. Based on GFRNet, Dogan \etal \cite{dogan2019exemplar} propose a GWAInet for face super-resolution, which is trained in an adversarial generative manner to generate high-quality face images. Compared with GFRNet, GWAInet does not rely on facial landmarks in the training stage, which guides the model to focus more on the whole face region and increases the robustness of the model. These two methods use a WarpNet to predict the flow field to warp the reference to align with the low-quality images. However, the alignment still does not fully solve all the differences between the reference and low-quality images, \ie mouth close to open. Furthermore, these two methods rely on a high-quality reference image with the same identity, which makes them only applicable in limited scenes. Thus, Li \etal \cite{li2020blind} propose DFDNet for face restoration. It uses deep component dictionaries as reference priors to support restoration. DFDNet first applies K-means to build facial component dictionaries for key components, including the left eye, right eye, nose, and mouth, from high-quality images. It then selects the most similar component features from these dictionaries and transfers the details to the low-quality face image to guide restoration. Recently, Li \etal \cite{li2022learning} propose a DMDNet model, which explicitly memories the generic and specific features through dual facial dictionaries in the network for blind face restoration. To recover face well on mobile phones with limited memory and processing budget, Lai \etal \cite{lai2022face} develop a face deblurring system based on the dual camera fusion technique for mobile phones, which can promote the further development of reference-based face deblurring.


\textbf{Generative Prior Based Deep Restoration Methods}. This group leverages pretrained generative models or latent priors to synthesize plausible facial details and improve perceptual realism under severe degradations. With the rapid development of generative adversarial network (GAN)~\cite{karras2019style,karras2020analyzing}, recent works find that~\cite{gu2020image,menon2020pulse} the generative prior of pre-trained face GAN models, such as StyleGAN~\cite{karras2019style} and StytleGAN2~\cite{karras2020analyzing}, can provide rich and diverse facial information (\eg geometry and facial textures). And researchers have started to leverage the GAN prior to face image restoration. The first kind of generative prior based methods is inspired by GAN inversion methods, which mainly aim at finding the closest latent vector in the GAN span from the input image. PULSE~\cite{menon2020pulse} is another representative method, which is a self-supervised face restoration method by optimizing the latent of a pre-trained StyleGAN~\cite{karras2019style}. Inspired by PULSE, mGANprior~\cite{gu2020image} considers multiple latent codes in the pre-trained GAN and optimizes multiple codes to promote the ability of image reconstruction. However, these methods fail to preserve fidelity in the restored face images. To achieve a better balance between visual quality and fidelity of the recovered images, recent works GFP-GAN~\cite{wang2021towards} and GPEN~\cite{yang2021gan} first extract fidelity information from the input low-quality face images and then leverage the pre-trained GAN as a decoder to capture the facial prior. Specifically, GFP-GAN~\cite{wang2021towards} uses the facial distribution learned by a pre-trained GAN as a facial prior to achieve joint restoration and color enhancement. It includes a degradation removal module and a pre-trained face GAN prior. These two modules are connected by latent code mapping and several channel-split spatial feature transform layers. This design helps balance realness and fidelity. GPEN~\cite{yang2021gan} aims to effectively integrate the advantages of GAN and DNN for face restoration. In GPEN, it first learns a GAN used for generating high-quality face images and embeds this pre-trained GAN into a deep neural network as a decoder prior for face restoration. More recently, Zhu \etal \cite{zhu2022blind} propose to combine shape and generative prior to guide the process of face restoration for the network. In the proposed network, they first use a shape restoration module to generate a shape prior. Then, a shape and generative prior integration module is proposed to fuse the shape and generative prior. Finally, they introduce a hybrid-level loss to jointly optimize the shape and generative prior with other network parts, and thus, these two priors can better benefit the face restoration. To address severe degradation in facial images, Xie \etal \cite{xie2023tfrgan} introduce a novel framework named TFRGAN. This framework focuses on generating a more accurate and improved latent code for the StyleGAN2 prior by incorporating both text and image information within the latent code space. In addition to utilizing generative priors, recent methods~\cite{hou2023semi,hu2023dear,cheikh2023degradation} also propose novel frameworks to leverage the representation of degradation and generative priors in face images. These methods aim to achieve a balance between realism and fidelity when dealing with diverse levels of degradation.

\textbf{Non-prior Based Deep Restoration Methods}. This group learns a direct mapping from low-quality to high-quality faces without explicit facial priors, often focusing on efficiency and general restoration robustness. Although most deep learning-based FR methods can recover satisfying faces with the help of the facial prior, it makes the cost of generating face images expensive and laborious. To address this problem, many methods aim to design a network that directly learns the mapping function between low-quality and high-quality face images without any additional facial priors. Some techniques are introduced in the models to improve the feature representation, such as multi-path structure, attention mechanism, feature fusion strategy, adversarial learning, strong backbone \etc 

The first representative work is dated back to 2015. Zhou \etal \cite{zhou2015learning} propose a bi-channel convolutional neural network (BCCNN) for face super-resolution. It consists of a feature extractor and an image generator. The proposed feature extractor extracts robust face representations from the low-resolution face image. The image generator is designed to adaptively fuse the extracted face representations and the input face image to generate a high-resolution image. BCCNN can achieve better restoration results for the face image with large variations. However, this work directly ignores pre-aligned facial spatial configurations (such as facial landmark localization) and thus does not perform well when the input image has severe blur. To address this problem, Zhu \etal \cite{zhu2016deep} propose a cascade bi-network called CBN to jointly optimize facial dense correspondence field estimation and face super-resolution. CBN obtains better performance results than previous works. However, when the face feature location in the model is wrong, CBN may generate ghosting face images.

Following previous works~\cite{zhou2015learning,zhu2016deep}, some state-of-the-art methods~\cite{cao2017attention,yu2017hallucinating,wang2017deepdeblur,jiang2020dual,wang2023spatial} focus on designing different CNN networks and learning strategies, such as partial~\cite{liu2018image} and Fourier~\cite{suvorov2022resolution} convolution, recurrent learning strategy~\cite{li2020recurrent}, and multi-path structure~\cite{jiang2020dual}, to improve the performance of the network. Among them, \cite{jiang2020dual} is a representative work that aims at using recurrent and multi-path structures in the network to improve performance. Jiang \etal \cite{jiang2020dual} propose a dual-path deep fusion network (DPDFN) for face super-resolution. The core insight of DPDFN is local and global feature learning and fusion in two branches. A convolutional denoising autoencoder network is proposed in~\cite{tun2020facial} for face denoising, which can achieve superior denoising performance by leveraging robust spatial correlations. Over the past few years, GAN~\cite{goodfellow2014generative} has become another popular technology in the computer vision community. It has been widely applied in many applications, including image synthesis, semantic image editing, style transfer, classification, and image restoration. 
Compared with CNN, GAN can generate more realistic images~\cite{creswell2018generative}. The typical GAN structure consists of a generator network and a discriminator network. The generator is designed to produce realistic images, and the discriminator is used to figure out the difference between the image produced by the generator and the real image. The generator and discriminator are trained at the same time and compete against each other. In 2016, Yu and Porikli~\cite{yu2016ultra} make the first attempt to develop GAN and propose ultra-resolution by discriminative generative networks (UR-DGN) for face restoration. In UR-DGN, through an adversarial learning strategy, the discriminant network is used to learn the important components of human faces, and the generation network fuses these facial components into the input image.
Following Yu and Porikli~\cite{yu2016ultra}, many GAN based face restoration methods are proposed in the literature~\cite{xu2017learning,yu2017face,bulat2018learn,shao2019potentials,shiri2019identity,yang2020hifacegan}. These methods integrate many techniques (\eg loss functions, learning strategies, identity constraints \textit{etc.}) into the GAN network and achieve better visual results. Specifically, MCGAN~\cite{xu2017learning} uses a multi-class GAN model and a feature matching loss. TDN~\cite{yu2017face} aims to exploit the class specific information in the process of restoration. HLGAN~\cite{bulat2018learn}, tfh-WGAN~\cite{shao2019potentials}, and HiFaceGAN~\cite{yang2020hifacegan} focus on designing more complex GAN models, including two-stage GANs, WGAN, and multi-stage GAN network. IFRP~\cite{shiri2019identity} adopts identity-preserving algorithms to help the GAN model produce high-quality face images with accurate identity information. 

\begin{figure*}[t]
\begin{center}
 \begin{overpic}[width=0.9\textwidth]{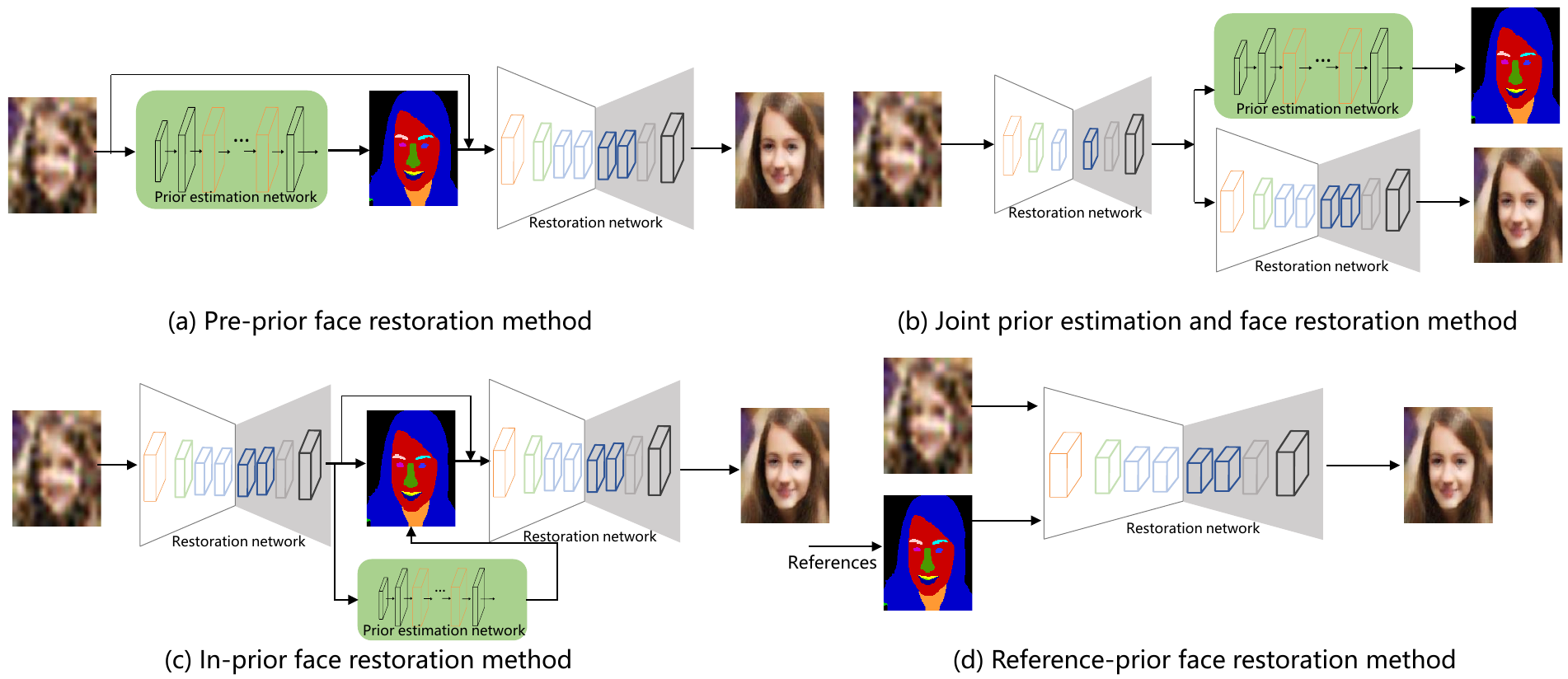}

 \end{overpic}
 \vspace{-3mm}
 	\caption{Summary of the network architecture of prior-guided methods. It mainly consists of Pre-prior face restoration methods, Joint prior estimation and face restoration methods, In-prior face restoration methods, and Reference-prior face restoration methods. We use the facial parsing map as an example of prior in the figure.}
	\label{fig:network_architectures}
\end{center}
	\vspace{-6mm}
\end{figure*}
Since 2014, the attention mechanism has been gradually applied to visual tasks and has achieved great effects~\cite{mnih2014recurrent,hu2018squeeze}. The core idea of the attention mechanism is to reweight features through a learnable weight map to emphasize the important features and suppress the less useful ones. Many face restoration methods~\cite{chen2020learning,wang2020face,zhao2020saan,lu2021face} resort to the attention mechanism to improve their performance. Among them, \cite{chen2020learning,wang2020face,zhao2020saan} mainly design large-scale residual blocks with the attention mechanism to extract fine-grained face features, which can produce better performance. However, they do not consider cross-channel interaction in the residual blocks, which reduces the ability of feature representation in the network. Thus, Lu \etal \cite{lu2021face} propose a split-attention in the split-attention network (SISN) for face super-resolution. SISN is stacked by several external-internal split attention group (ESAG) modules. ESAG uses multi-path learning, attention mechanism, and residual learning to enable the network to focus on facial texture details and structure information simultaneously. With this specific module, SISN can generate high-quality faces containing more facial structural information. In addition, some recent works~\cite{yu2018generative,zeng2023self,tomar2023attentive,tomar2023deep} also design different attention mechanisms in the network to enhance the visual quality of the generated face images. For example, \cite{zeng2023self} proposes a self-attention learning network that utilizes a complementary three-stage face super-resolution architecture and a simple self-attention module to enhance the degraded input face. In \cite{tomar2023attentive,tomar2023deep}, the authors use spatial attention to help the network focus on preserving the facial structure features. In recent years, the Vision Transformer (ViT) architecture has shown great potential in computer vision. Many methods~\cite{wang2022restoreformer,li2022faceformer,zhou2022towards,zhang2022blind,gao2023ctcnet,bao2023sctanet,shi2023exploiting,wei2023composite} aim to use strong Transformer backbone to build face restoration networks and have demonstrated superior performance. Among them, several studies~\cite{wang2022restoreformer,li2022faceformer,zhou2022towards,zhang2022blind} directly resort to the Transformer to enhance the network's global representation ability, resulting in improved performance. However, relying solely on image-level self-attention might lead to the loss of local fine-grained details. Therefore, some studies~\cite{li2022mat,bao2023sctanet,qi2023efficient,shi2023exploiting,wei2023composite} aim to combine the strengths of convolutional neural networks (CNN) and Transformers to effectively utilize both global information and local features, thereby achieving high-quality face reconstruction. For example, Bao \etal \cite{bao2023sctanet} propose a spatial attention-guided CNN-Transformer aggregation network (SCTANet) for FSR. The core components are the hybrid attention aggregation block and the sub-pixel MLP-based upsampling module.  In~\cite{qi2023efficient},~\cite{shi2023exploiting} and \cite{wei2023composite}, the authors also propose FSR frameworks incorporating Transformer and CNN architectures known as ELSFace, DBNet, and MOHA, respectively. Recent works also explore more efficient architectures and stronger multiscale feature modeling for face super-resolution. WFEN proposes a wavelet-based feature enhancement network to better preserve high-frequency facial details while maintaining efficiency~\cite{li2024efficient}.
AMINet introduces an attention-guided multiscale interaction design to improve feature fusion and complementarity in hybrid architectures~\cite{wan2025attention}. In addition, diffusion models, such as the diffusion denoising diffusion probability models (DDPM)~\cite{ho2020denoising} and the denoising diffusion implicit models (DDIM)~\cite{song2020denoising}, have garnered considerable attention in the task of face restoration~\cite{gao2023multi,dos2022face,nair2023ddpm,qiu2023diffbfr,wang2023dr2,yue2022difface,zhao2023towards}. When compared to other models, diffusion models exhibit greater capability in representing image pixel distribution. This characteristic presents substantial potential for enhancing visual quality and benefiting high-quality face restoration. For example, Wang \etal \cite{wang2023dr2} propose a diffusion-based robust degradation remover called DR2 for face restoration. DR2  involves initially transforming the degraded image into a coarse, degradation-invariant prediction. Subsequently, an enhancement module is employed to restore the coarse prediction and generate a high-quality face image. Gao \etal \cite{gao2023multi} propose a novel conditional generative model called BPSR3 for face super-resolution, which is based on diffusion models. Specifically, BPSR3 replaces the original U-Net, which is used in super-resolution via repeated refinement (SR3), with a multi-scale deep back-projection network structure. SSDiff proposes a self-supervised selective-guided diffusion framework for old-photo face restoration by using region guidance such as parsing maps and scratch masks~\cite{li2025ssdiff}. MCS introduces measurement-constrained sampling to enable diverse prompt-aligned reconstructions under severe degradations~\cite{li2025measurement}.

Although face restoration is mainly studied as a low-level vision problem, it is also widely used as a preprocessing step for downstream face analysis tasks under real-world degradations. Typical applications include face detection and tracking in low-resolution, blurred, or compressed imagery, face recognition/verification with identity preservation requirements, as well as video analytics and forensic analysis in unconstrained scenarios~\cite{ataer2018super,singh2022towards,wang2023efficientsrface,yang2024effects}. In these settings, restoration can potentially improve the robustness of subsequent models by enhancing facial structures and removing degradations. However, it may also introduce non-authentic details (``hallucination'') that improve perceptual quality but negatively affect the reliability of downstream decisions, especially for detection/recognition and forensic use cases~\cite{wang2023efficientsrface,singh2022towards}. The above discussion methods focus on single face images. Recently, a few works have started to explore video face restoration by explicitly modeling temporal information across frames. For example,  Chen \etal \cite{chen2024towards} introduce a real-world video face restoration benchmark to support evaluation under practical degradations. Xu \etal \cite{xu2024beyond} propose PGTFormer, which uses parsing-guided temporal-coherent transformer modeling to better preserve facial structures and improve temporal consistency in blind video face restoration. Wang \etal \cite{wang2025svfr} propose SVFR, a unified framework for generalized video face restoration that improves robustness across different degradations and scenes. In general, these methods combine spatial restoration with temporal modeling, such as temporal feature aggregation, temporal attention, and temporal consistency constraints, to reduce flickering artifacts and stabilize identity related details across frames. Video face restoration remains challenging under large motion, occlusion, and mixed degradations, and further progress is needed on stronger temporal modeling and efficient inference.

\section{Technical Development Review}\label{sec:development_review}
In this section, we discuss the developments of existing face restoration in the following aspects: network architectures, basic blocks, loss functions, and benchmark datasets. 

\subsection{Network Architectures} 

Existing state-of-the-art networks are designed by focusing on facial prior, pre-trained GAN models, ViT architectures, and diffusion models. Thus, we discuss these developments in this section.  

\textbf{Prior-guided Networks}. As a domain-specific image restoration task, it is important to consider the characteristics of face images (\eg identity, structure, and face pose) when designing the specialized networks for face restoration. To this end, some priors are introduced into the networks to help the process of restoration. With the help of facial priors, these networks can generate realistic faces with details. According to the way of using priors, the architectures of the prior-guided networks can be divided into four categories: Pre-prior face restoration method, Joint prior estimation and face restoration method, In-prior face restoration method, and Reference-prior face restoration method. The summary of these architectures is illustrated in Fig.~\ref{fig:network_architectures}.


\begin{figure*}[t]
\begin{center}
 \begin{overpic}[width=0.8\textwidth]{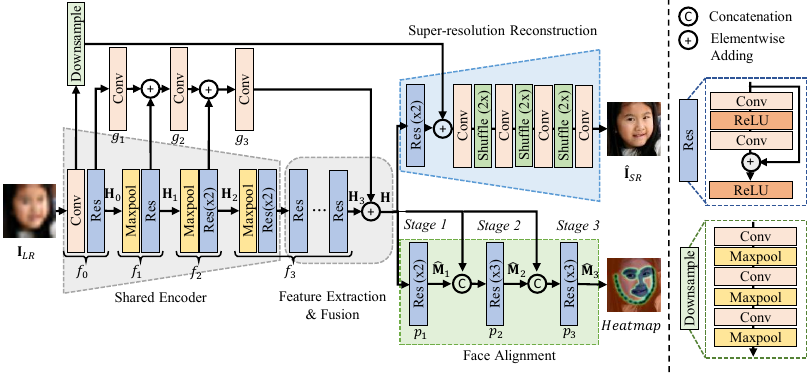}
 \end{overpic}
  \vspace{-3mm}
 	\caption{Architecture to jointly learn landmark localization and face restoration in~\cite{yin2020joint}.} 
	\label{fig:architectures_examples3}
\end{center}
	\vspace{-6mm}
\end{figure*}


For pre-prior face restoration methods~\cite{shen2018deep,shen2020exploiting,kalarot2020component,chen2021progressive}, they usually adopt a prior estimation network (\eg face passing network or a pre-trained face GAN) to extract prior from the low-quality input. For example, ~\cite{shen2018deep} designs a face parsing network to extract the semantic label from the input image or coarse deblurred image. Then it concatenates the input blurred image and the face semantic label to the deblurring network to generate the sharp image.

The second type of method is the joint prior estimation and face restoration method, which takes advantage of the relationship between the prior estimation task and the face restoration task. These methods \cite{zhu2016deep,li2018face,li2020learning,yin2020joint} usually jointly train the face restoration network and the prior estimation network. This kind of method enjoys the benefit of two sub-tasks and directly promotes face restoration performance. For example, as illustrated in Fig. \ref{fig:architectures_examples3}, 
Yin \etal \cite{yin2020joint} propose a joint alignment and face super-resolution network to jointly estimate facial landmarks and super-resolve face images.

\begin{figure*}[t]
\begin{center}
 \begin{overpic}[width=0.8\textwidth]{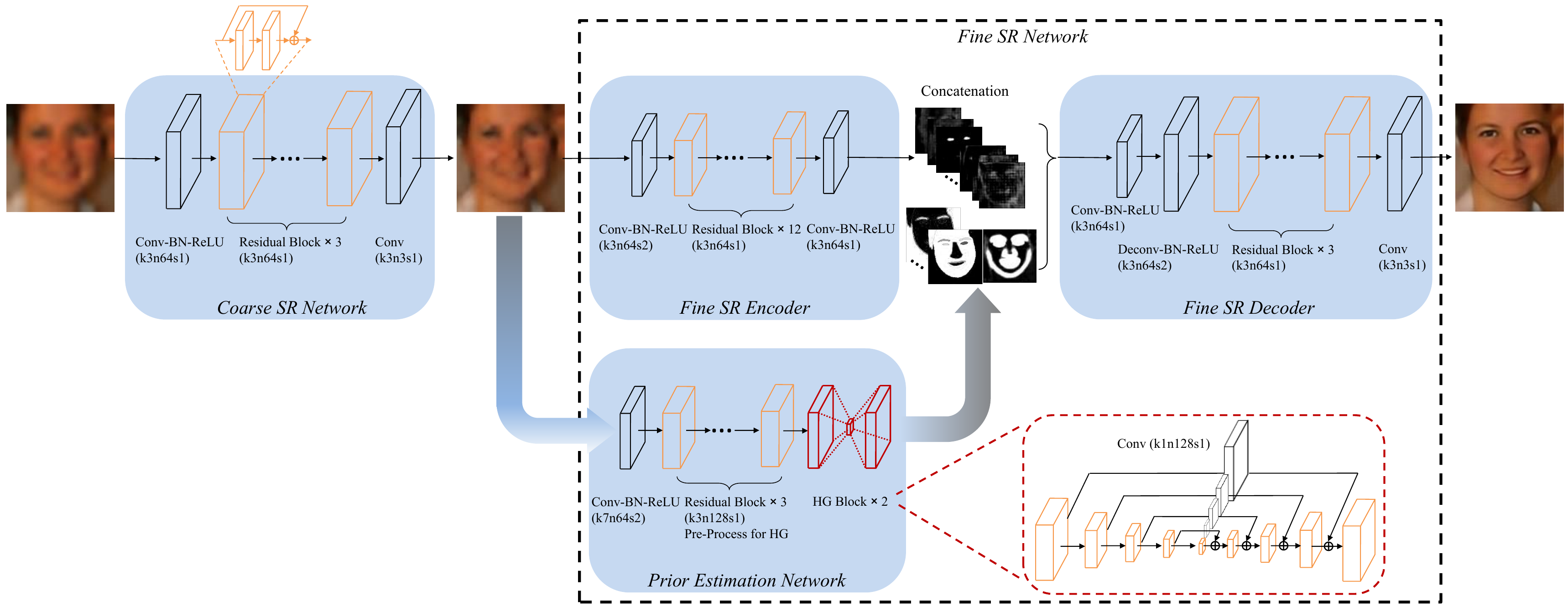}
 \end{overpic}
\vspace{-3mm}
 	\caption{Architecture to estimate face passing map in the middle of the network in \cite{chen2018fsrnet}.} 
	\label{fig:architectures_examples2}
\end{center}
	\vspace{-6mm}
\end{figure*}

However, directly extracting the face prior from low-quality images is difficult. Thus, in-prior face restoration methods~\cite{chen2018fsrnet,yu2018face,shen2020exploiting,ma2020deep} first use a restoration network to produce the coarse recovered image, then extract the prior information from the coarse image, which can obtain a more accurate prior. FSRNet~\cite{chen2018fsrnet} is one representative method, which is shown in Fig. \ref{fig:architectures_examples2}. In FSRNet, a coarse SR network is used to recover the coarse image, and then the coarse image with high quality is processed by a fine SR encoder and a prior estimation network respectively. After that, both image features and prior information are fed to the fine SR decoder to recover the final results. 

In contrast to the above methods that estimate face priors directly or indirectly from low-quality images, reference-prior face restoration methods aim to exploit the high-quality images of the same person to alleviate the difficulty of facial prior estimation or image restoration. Some methods~\cite{li2018learning,dogan2019exemplar} propose a warping subnet to align the reference and degraded images. Typical works are GFRNet~\cite{li2018learning} and GWANet~\cite{dogan2019exemplar}. In GFRNet~\cite{li2018learning}, a landmark loss and a total variation regularization are designed to train the warping subnet. GWANet~\cite{dogan2019exemplar} trains the warping subnet in an end-to-end manner without the facial landmark and proposes a feature fusion chain with multiple convolution layers to fuse features from the warped guidance and degraded image. Recent works~\cite{li2020blind,li2020enhanced} propose to exploit deep facial component dictionaries or use multiple high-quality exemplars in the face restoration to exploit more guidance features and thus improve the generalization ability when dealing with low-quality face images with unknown degradation.

\begin{figure*}[t]
\begin{center}
 \begin{overpic}[width=\textwidth]{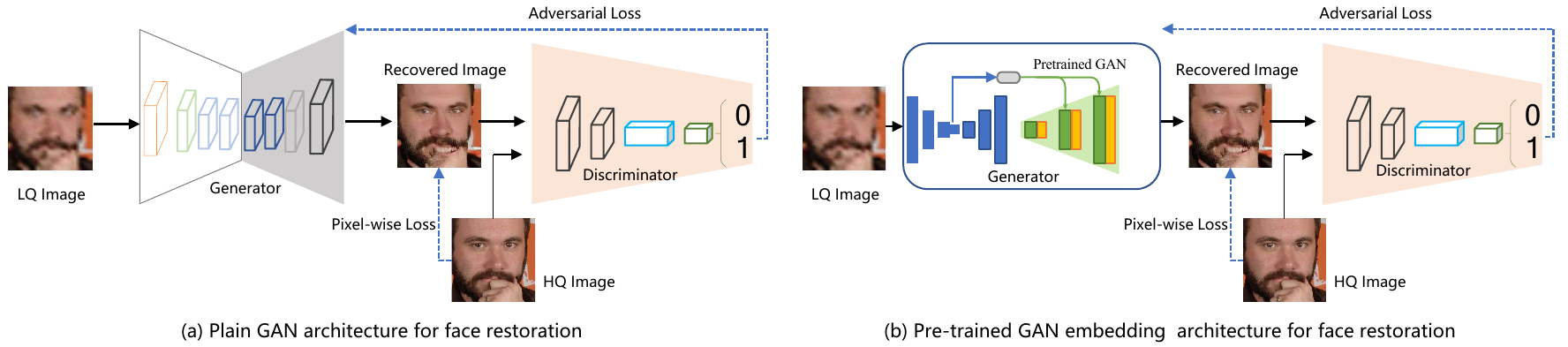}
 \end{overpic}
 \vspace{-3mm}
 	\caption{Summary of GAN architecture used for face restoration. It mainly contains a plain GAN architecture and a pre-trained GAN embedding architecture.} 
	\label{fig:network_architectures_GAN}
\end{center}
	 \vspace{-5mm}
\end{figure*}

\textbf{GAN-based Networks}. With the success of the GAN architecture, some works aim at designing specialized GAN networks for face restoration. As shown in Fig.~\ref{fig:network_architectures_GAN}, the architectures of GAN-based networks can be summarized as the plain GAN architecture and the pre-trained embedding architecture. In the plain GAN architecture-based methods~\cite{yu2016ultra,xu2017learning,yu2017face,bulat2018learn,yang2020hifacegan}, they introduce an adversarial loss in the network and use adversarial learning to jointly optimize the discriminator and generator (\ie face restoration network) to generate realistic face images. Among them, HLGAN~\cite{bulat2018learn} is one representative method for face super-resolution. HLGAN consists of two generative adversarial networks. The first network is a High-to-Low GAN, which is trained with unpaired images to learn the degradation process of the high-resolution images. After that, its outputs (\ie low-resolution face images) are adopted to train a Low-to-High GAN for face super-resolution. The second Low-to-High GAN is trained with paired face images. Thanks to this two-stage GAN architecture, HLGAN can achieve superior performance when dealing with real face images.

\begin{figure*}[t]
\begin{center}
 \begin{overpic}[width=0.8\textwidth]{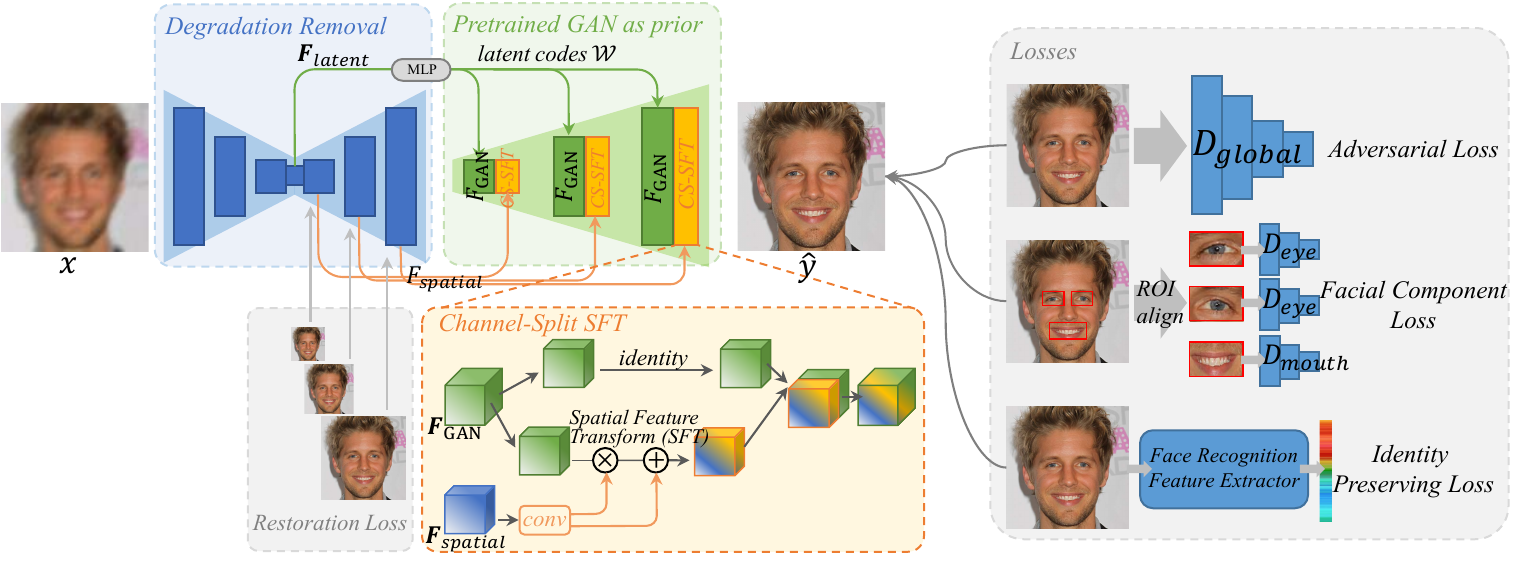}
 \end{overpic}
\vspace{-3mm}
 	\caption{The architecture of GFP-GAN~\cite{wang2021towards}, which is designed under the pre-trained GAN embedding architecture.} 
	\label{fig:GFP_GAN}
\end{center}
    \vspace{-6mm}
\end{figure*}

In pre-trained GAN embedding architecture-based methods~\cite{gu2020image,menon2020pulse,wang2021towards,yang2021gan,zhu2022blind}, they 
exploit the latent prior in pre-trained face GAN models such as StyleGAN~\cite{karras2019style} and incorporate the prior into the process of face restoration. One representative work is GFP-GAN, which effectively leverages face priors encapsulated in the pre-trained face GAN to perform face restoration.
The detail architecture of GFP-GAN~\cite{wang2021towards} is illustrated in Fig.~\ref{fig:GFP_GAN}. Specifically, GFP-GAN is composed of a degradation removal module and a pre-trained face GAN. These modules are connected together by the latent code mapping and some channel-split spatial feature transform layers.
In addition, a loss function combined with the pixel-wise loss, the facial component loss, the adversarial loss, and the identity preserving loss is proposed to train the GFP-GAN. With these techniques, GFP-GAN can recover high-quality face images with facial details.


\textbf{ViT-based Networks}. In recent years, the Visual Transformer (ViT)~\cite{dosovitskiy2020image} architecture has demonstrated superior performance in natural language processing and computer vision. ViT triggers the direct application of the Transformer architecture~\cite{vaswani2017attention} in computer vision tasks, including object recognition, detection, and classification~\cite{dosovitskiy2020image,carion2020end,wang2021pyramid}. ViT architecture also begins to be applied to the face restoration task. Wang \etal \cite{wang2022restoreformer} propose RestorFormer based on ViT architecture for face restoration. RestorFormer aims at modeling contextual information of the face image to help the process of face restoration. Specifically, Wang \etal propose a novel multi-head cross-attention layer, which explores spatial interactions between corrupted queries and high-quality key-value pairs. The high-quality key-value pairs are from a learned high-quality dictionary. 
With the help of advanced architecture and the high-quality dictionary prior, RestorFormer recovers results with more texture details and complete structures. Zhou \etal \cite{zhou2022towards} treat face restoration as a code prediction task that aims at learning the discrete codebook prior in a small finite proxy space. They thus propose a Transformer-based prediction network (CodeFormer) to achieve code prediction for face restoration.

To effectively process diverse scale face images, Li \etal \cite{li2022faceformer} propose a novel scale-aware blind face restoration framework FaceFormer. They transform facial feature restoration as a facial scale-aware transformation procedure and then employ hierarchical Transformer blocks in the network to extract robust facial features. The FaceFormer produces high-quality face images with faithful details. Inspired by the success of Swin Transformer~\cite{liu2021swin} on high-level vision tasks, Zhang \etal \cite{zhang2022blind} propose an end-to-end Swin Transformer U-Net (STUNet) for face restoration. In STUNet, the self-attention mechanism and the shifted windowing scheme are used to help the model focus on important features for effective face restoration. Besides, they build two larger-scale face restoration benchmark datasets to further advance the development of face restoration and model evaluation. Although the above ViT-based methods have demonstrated effectiveness in face restoration, there are still many problems to be studied, such as the model's efficiency and generalization in real-world scenery.

\textbf{Diffusion Models based Networks}. Recently, diffusion probabilistic models~\cite{ho2020denoising,sohl2015deep,song2020denoising} have shown state-of-the-art performance in computer vision tasks. The core technique of diffusion probabilistic models is transforming the complex and unstable generation process into multiple independent and stable reverse processes through the use of Markov Chain modeling. Diffusion models have delivered strong performance across multiple areas, such as data synthesis, image understanding, and low-level vision. For low-level vision in particular, prior diffusion-based approaches~\cite{gao2023implicit,ren2023multiscale,ozdenizci2023restoring,he2025diffusion} are mainly designed to recover images from degraded observations, producing visually faithful reconstructions with clear semantic structure and fine-grained, realistic textures even when the degradation is severe and highly complex. In the task of face restoration, the denoising diffusion probabilistic model (DDPM)~\cite{ho2020denoising} is early utilized. DDPN contains two key processes in the diffusion model, namely the forward process and the reverse process, which are briefly introduced as follows.

The forward process (\ie the diffusion process) is a fixed Markov Chain that sequentially corrupts an image $x_0 \sim p_{data}(x)$ at $T$ diffusion time steps, by injecting Gaussian noise according to a variance schedule $\beta_{1},...,\beta_{T}\in(0,1)$. It can be formulated as:
\begin{equation}
    q(x_{t}|x_{t-1})=\mathcal{N}(x_{t};\sqrt{1-\beta_{t}} \cdot x_{t-1},\beta_{t}\mathbf{I}).
    \label{eq:diffusion_process}
\end{equation}
 Moreover, we can compute the probabilistic distribution of $x_t$ given $x_0$ as:
\begin{equation}
    q(x_{t}|x_{0})=\mathcal{N}(x_{t};\sqrt{\hat{\alpha}_{t}} x_{0},\sqrt{1-\hat{\alpha}_{t}}\mathbf{I}), 
\end{equation}
where $\alpha_{t}=1-\beta_{t}$ and $\hat{\alpha}_{t}=\prod \limits_{i=1}^t \alpha_{i}$. Then, $x_{T}\sim\mathcal{N}(0, \mathbf{I})$ if $T$ is larger enough. 

\begin{figure}[t]
	\centering
	\includegraphics[width=0.8\linewidth]{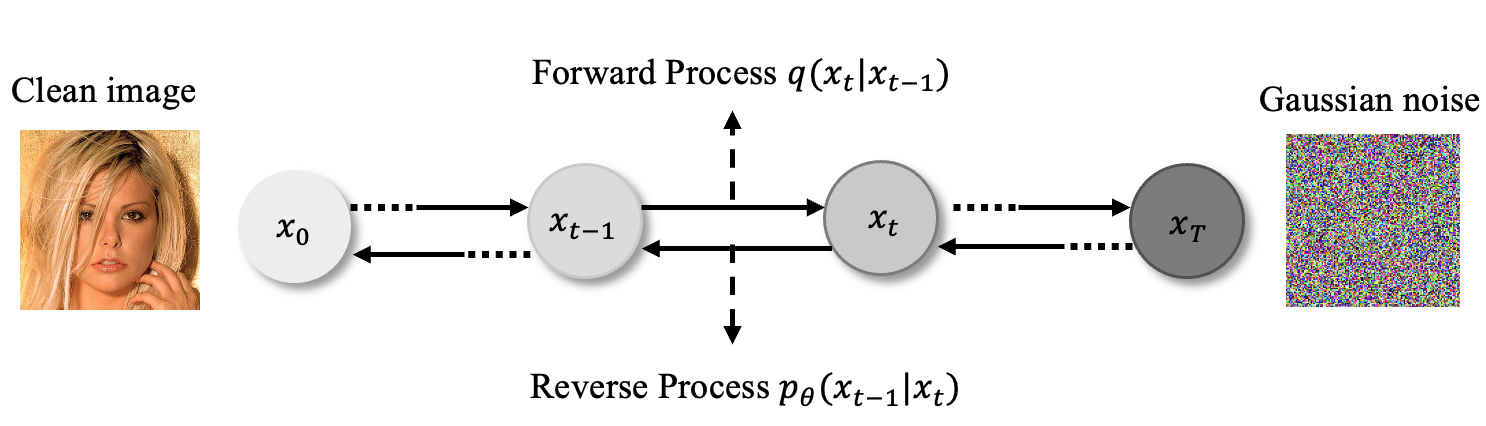}
    \vspace{-3mm}
	\caption{Illustration of denoising diffusion probabilistic models.}
	\label{fig:DDPM}
    \vspace{-6mm}
\end{figure}

The forward process aims to generate images in a progressive manner, which is achieved through a Gaussian transition with a learned mean $\mu_\theta$:
\begin{equation}
    p_\theta(x_{t-1}|x_{t}) = \mathcal{N}(x_{t-1}; \mu_\theta (x_t, t),\Tilde{\beta}_t\mathbf{I}), 
    \label{eq:reverse}
\end{equation}
where $\mu_\theta (x_t, t)=\frac{1}{\sqrt{\alpha_t}}(x_t-\frac{\beta_t}{\sqrt{1-\hat{\alpha}_t}})\epsilon_\theta (x_t, t)$ for $\epsilon \sim \mathcal{N}(\mathbf{0}, \textbf{I})$  and $\tilde{\beta_t}=\frac{1-\hat{\alpha}_{t-1}}{1-\hat{\alpha}_t}$. The variance schedule $\beta_t$ is predefined, and thus, it only requires approximating the mean $\mu_\theta (x_t, t)$ by a denoising network $\epsilon_\theta (x_t, t)$. Fig.~\ref{fig:DDPM} shows the forward and reverse processes of denoising diffusion probabilistic models.


Diffusion based face restoration typically performs restoration through an iterative denoising process. A forward process gradually corrupts a clean image with noise, and a learned reverse process removes noise step by step to recover a high quality face. Compared with feed-forward restoration networks that produce the output in a single pass, iterative denoising can better align the output distribution with natural face images, which often improves perceptual quality. Thus, recent researchers utilize diffusing models for face restoration~\cite{wang2025osdface,liu2025faceme,lu2025visual}. Different from existing diffusion-based methods (\eg SR3~\cite{saharia2022image}) that use U-Net as their backbone network, Gao \etal \cite{gao2023multi} replace the U-Net in super-resolution via repeated refinement (SR3) with a multi-scale deep back-projection network structure. This modified diffusion model obtains a faster convergence and better restoration quality for face images with fewer parameters than the original SR3. In~\cite{yue2022difface}, the authors introduce a diffusion-based method named DifFace for face restoration. Notably, DifFace utilizes a pre-trained restoration network, such as SRCNN~\cite{dong2015image} or SwinIR~\cite{liang2021swinir}, to acquire an initial clean image that serves as the starting point for the diffusion model's sampling process. This unique design enhances the generalization capability of the diffusion model in handling severe and unknown degradations in face images. In contrast, Wang \etal \cite{wang2023dr2} propose a diffusion-based framework called DR2E for face restoration. Specifically, DR2E contains two stages. In the first stage, DR2 transforms the degraded images into coarse results that belong to a degradation-invariant distribution. In the second stage, the enhancement module further processes the degradation-invariant images to produce high-quality details. Diffusion-based designs are attractive because iterative denoising can better align outputs with the natural face distribution and often yields strong perceptual realism under severe degradations. They can also incorporate conditioning signals such as degraded inputs and facial priors to improve controllability and robustness. However, diffusion models usually require multi-step sampling, which increases inference time. Thus, an important future direction is to develop faster sampling and distilled diffusion strategies, and to improve identity-consistent and attribute-controllable restoration for practical deployment.

\subsection{Basic Blocks}
In the field of face restoration, different types of basic blocks are designed to build powerful face restoration networks. In this section, we detail basic blocks that are widely used in the network.

\textbf{Residual Learning and Dense Connections}. Residual learning and dense connections are two widely adopted strategies in face restoration to enable deeper networks, mitigate gradient issues, and promote feature reuse.

\textit{Residual Learning}. Introduced in ResNet by He \etal~\cite{he2016deep}, residual blocks use skip connections to alleviate vanishing/exploding gradients and allow deeper architectures. Many face restoration methods~\cite{chen2018fsrnet,kim2019progressive,hu2021face,chen2021progressive} adopt residual learning, which can be categorized into two types:

\begin{itemize}
    \item Global Residual Learning (GRL) focuses on predicting the residual between the input and target images rather than generating the clean image directly~\cite{tai2017image}. Since most residual values are near zero, this simplifies learning and improves convergence~\cite{chen2021progressive,yang2020hifacegan,zhang2022blind}.
    
    \item Local Residual Learning (LRL) applies residual connections within network modules. It is frequently used to stabilize training and enhance feature extraction~\cite{wang2017deepdeblur,bulat2018learn,wang2021towards}. For instance, Chen \etal~\cite{chen2018fsrnet} utilize residual blocks in FSRNet to construct coarse/fine SR networks and a prior estimation module. Some works combine GRL and LRL to leverage the benefits of both~\cite{yang2020hifacegan,chudasama2021comsupresnet,zhang2022blind}.
\end{itemize}

\textit{Dense Connections}. Huang \etal~\cite{huang2017densely} propose DenseNet, where each layer connects to all preceding layers. This architecture encourages feature reuse, strengthens gradient flow, and supports multi-level feature fusion. In face restoration, dense connections have proven effective~\cite{tu2021joint,jiang2020dual,liu2020densely}. For example, Tu \etal~\cite{tu2021joint} apply dense links in the decoder to enhance spatial detail reconstruction. Jiang \etal~\cite{jiang2020dual} further integrate residual, dense, and recursive modules in DPDFN, forming a memory-driven sub-network that improves information flow and boosts performance.

\textbf{Attention Mechanism}. With the successes of the attention mechanism in various vision tasks, the attention mechanism has been widely employed in face restoration networks~\cite{hu2021face,cao2017attention,kim2019progressive}. Among them, the utilized attention techniques can be divided into channel attention, spatial attention, hybrid attention, and other attention mechanisms.

\textit{Channel Attention}. The channel attention is to learn the relative weights between feature channels and make the model focus on the important feature channels. For example, Chudasama \etal \cite{chudasama2021comsupresnet} propose an E-ComSupResNet network with channel attention to super-resolve low-resolution face images. In E-ComSupResNet, it integrates the channel attention into the Resblock to rescale the channel-wise feature maps adaptively.

\textit{Spatial Attention}. The spatial attention focuses on capturing the spatial contextual information of the feature. For instance, Chen \etal \cite{chen2020learning} introduce a
spatial attention mechanism to the residual blocks and use the modified block to build a network. 
With the guidance of spatial attention, the network can pay more attention to features related to the key face structures. 

\textit{Hybrid Attention}. Some methods use channel and spatial attention mechanisms to improve the representation of the network. For instance, to exploit the 3D face rendered priors in the network, Hu \etal \cite{hu2021face} develop a spatial attention module (SAM) with channel and spatial mechanisms to capture the locations of face components and the facial identity. This module effectively exploits the hierarchical information of 3D faces to help the network generate high-quality face images.

\textit{Other Attention}. Some methods do not use the attention mechanism in the network design. On the contrary, they aim to propose an attention-based loss to optimize the network. One representative work is PFSN~\cite{kim2019progressive}, which uses facial heatmaps to produce a mask and obtains facial attention loss by computing the difference between the mask recovered and high-quality face images.    

\textbf{Transformer Block}. Due to its strong capability to capture long-range dependencies between sequences, the recent Transformer has been a popular architecture in the computer vision community. The vision Transformer architecture usually decomposes an input image into a sequence of local windows and uses the self-attention mechanism to learn their relationships. We divide the Transformer block into the plain Transformer and the Swin Transformer. 

\textit{Plain Transformer}. The original Transformer block is proposed in ~\cite{dosovitskiy2020image}. It contains a normalization layer, a multi-head self-attention layer, and a feed-forward network layer. Recent methods~\cite{wang2022restoreformer,zhou2022towards} employ the plain Transformer block to build network modeling global interrelations. For example, RestoreFormer~\cite{wang2022restoreformer} performs the cross-self-attention mechanism between corrupted queries (extracted from input image) and high-quality key-value (sampled from high-quality dictionary) pairs by Transformer blocks. With the help of Transformer blocks, RestoreFormer can recover a clear face with realness and fidelity.

\textit{Swin Transformer}. To reduce the complexity of the plain Transformer, Liu \etal \cite{liu2021swin} propose the Swin Transformer layer to build an efficient Transformer network called Swin Transformer. The main difference between the Swin Transformer layer and the plain Transformer is that it adopts local attention and window shifting mechanisms to realize multi-head self-attention. Due to its impressive performance, it has been used in the face restoration methods~\cite{li2022faceformer,zhang2022blind}. Li \etal \cite{li2022faceformer} use the Swin Transformer blocks in the network to effectively extract latent facial features. Zhang \etal \cite{zhang2022blind} integrate the Swin Transformer block into the UNet network to learn hierarchical facial features, achieving state-of-the-art performance in the face restoration task.

\subsection{Loss Functions}

To optimize face restoration networks, numerous loss functions have been proposed in the literature. 
In general, loss functions used in the existing methods can be approximately divided into pixel-wise loss, perceptual loss, adversarial loss, and face-specific loss. We review these representative loss functions in the following.

\textit{Pixel-wise Loss}. The pixel-wise loss measures the pixel-wise difference between the recovered image and its corresponding clear image. It can quickly match the feature distribution of restored and clear images and speed up network training. In existing methods, \text{L1} and \text{L2} losses are two widely-used pixel-wise losses in face restoration. They can be formulated as:
\begin{align}
	\label{eq:pixel-wise-loss}
	\mathcal{L}_{\text{1}}  & = \frac{1}{C W H} \sum_{c=1}^{C} \sum_{x=1}^{W} \sum_{y=1}^{H}\|{I_{hq(x, y,c)}}-{\hat{I}_{hq(x, y, c)}}\|_{1}, \\
	\mathcal{L}_{\text{2}} & = \frac{1}{C W H} \sum_{c=1}^{C} \sum_{x=1}^{W} \sum_{y=1}^{H}\|{I_{hq(x, y, c)}}-{\hat{I}_{hq(x, y, c)}}\|_2^2,
\end{align}	
where $I_{hq}$ and $\hat{I}_{hq}$ represent the ground-truth and recovered face images respectively. $W$ and $H$ denote the size of the image. $C$ refers to the channel of the image. It can be seen that \text{L2} loss is only sensitive to large errors, while \text{L1} loss treats larger and smaller errors equally. Early methods~\cite{chen2018fsrnet,bulat2018super,kim2019progressive} usually use \text{L2} loss in their models and recent works~\cite{yu2022multi,zhu2022blind,wang2022restoreformer} mainly resort to \text{L1} loss. While the pixel-wise loss can force the model to achieve high PSNR values, it often results in over-smooth and unrealistic images ~\cite{bulat2018super,zhang2022deep}. 

\textit{Perceptual Loss}. To generate more high-quality face images, methods~\cite{bulat2018super,kim2019progressive,yasarla2020deblurring,zhou2022towards} adopt a perceptual loss to train the network. The perceptual loss~\cite{johnson2016perceptual} computes the difference between the recovered image and the ground-truth image in the feature space of a pre-trained deep network such as VGG16, VGG19~\cite{simonyan2014very}, and ResNet~\cite{he2016deep}. The perceptual loss over a deep pre-trained network features at a given layer $i$ is shown as:
\begin{align}
\mathcal{L}_{\text{per}} & = \frac{1}{C_{i} W_{i} H_{i}} \sum_{c=1}^{C_{i}} \sum_{x=1}^{W_{i}} \sum_{y= 1}^{H_{i}}\left(\phi_{i}\left(I_{hq}\right)-\phi_{i}\left(\hat{I}_{hq}\right)\right)^{2},
\end{align}
where $\phi_{i}$ denotes the feature map obtained in the $i$-th layer of the pre-trained network, and $W_{i}$, $H_{i}$ represent the shape of the feature map. $C_{i}$ is the channel number. Benefiting from the perceptual loss, face restoration methods~\cite{bulat2018super,gu2020image,yu2022multi} generate visually-pleasing results.  

\textit{Adversarial Loss}. The objective of GAN-based face restoration methods~\cite{wang2021towards,yang2021gan,zhu2022blind} is based on the min-max game. The core idea is to learn a generator $\mathcal{G}$ to generate a high-quality face image such that the discriminator $\mathcal{D}$ cannot distinguish between the recovered image and the ground-truth image. This process can be expressed as solving the following min-max problem: 
\begin{align}
\min _{\mathcal{G}} \max _{\mathcal{D}} V(\mathcal{G}, \mathcal{D}) & = \mathrm{E}_{I_{hq} \sim p_{\operatorname{train}(I_{hq})}}[\log (\mathcal{D}(I_{hq}))] + \notag \\
&\mathrm{E}_{I_{lq} \sim p_{\mathcal{G}\left(I_{lq}\right)}}\left[\log \left(1-\mathcal{D}\left(\mathcal{G}\left(I_{lq}\right)\right)\right)\right],
\end{align}
where $I_{hq}$ and $I_{lq}$ are the high-quality face image and low-quality input image.
The adversarial loss from the discriminator to optimize the generator is formulated as:
\begin{align}
\mathcal{L}_{adv} & = \log \left(1-\mathcal{D}\left(\mathcal{G}\left(I_{lq}\right)\right)\right),
\end{align}
where $\mathcal{D}(\mathcal{G}(I_{lq})$ is the probability that the restored image is close to the ground truth image. With the help of adversarial loss, existing face restoration methods~\cite{yang2020hifacegan,wang2021towards,yang2021gan,zhu2022blind} can generate realistic textures in the recovered face image. 

\textit{Face-specific Loss}. As a highly structured object, the human face has its own special characteristics, thus some face-related losses are used in face restoration. This kind of loss aims at incorporating information related to the structure of the human face into the face restoration process. The widely-used one is heatmap loss~\cite{bulat2018super,kim2019progressive}, which is defined as:
\begin{align}
\mathcal{L}_{\text{heatmap}} & = \frac{1}{r^{2} N W H} \sum_{n=1}^{N} \sum_{x=1}^{r W} \sum_{y=1}^{r H}\left(M_{x, y}^{n}-\tilde{M}_{x, y}^{n}\right)^{2}, 
\end{align}
where $N$ represents the number of landmarks, $M$ and $\tilde{M}$ are face heatmaps that are calculated from the ground-truths and restored images respectively. Some works introduce human identity loss in the model. The identity preserving loss~\cite{wang2021towards} is shown as:
\begin{align}
\mathcal{L}_{id}  = \|\eta(\hat{I}_{hq})-\eta(I_{hq})\|_{1},
\end{align}
where $\eta$ is a face feature extractor, \eg ArcFace~\cite{deng2019arcface}, which is used to capture features for identity discrimination. In addition, many other face-specific loss functions are proposed, including facial attention loss~\cite{kim2019progressive}, face rendering loss~\cite{hu2021face}, semantic-aware style loss~\cite{chen2021progressive}, landmark loss~\cite{li2018learning}, facial component loss~\cite{wang2021towards}, and parsing loss~\cite{shen2020exploiting}. 
\begin{table*}[t]
	\centering
	\caption{Summary of benchmark datasets used in existing face restoration methods. $-$ indicates that the resolution of the image is not fixed. HQ-LQ represents the pairs of low-quality and high-quality face images in the dataset.}
    \vspace{-3mm}
	\label{tab:summary_dataset}
	\resizebox{0.7\linewidth}{!}{\begin{tabular}{c|c|cccc|c|c}
		\hline
		
		\multirow{2}{*}{Dataset} & \multirow{2}{*}{Size} & \multicolumn{4}{c|}{Additional Label} & \multirow{2}{*}{Resolution} & \multirow{2}{*}{HQ-LQ}\\
        	\cline{3-6}
         & & Attributes & Landmarks & Parsing maps &  Identity & &  \\
		\hline
		\hline
		\textbf{BioID}~\cite{jesorsky2001robust}    & $1,521$  & \XSolidBrush & $20$ & \XSolidBrush  & \XSolidBrush & $384 \times 286$ & \XSolidBrush   \\
		\hline
		\textbf{LFW}~\cite{huang2008labeled}   & $13,233$ & $73$ & \XSolidBrush &\XSolidBrush & \CheckmarkBold & $250 \times 250$ & \XSolidBrush   \\
		\hline
		\textbf{AFLW}~\cite{koestinger2011annotated}    & $25,993$ & \XSolidBrush & $21$ & \XSolidBrush & \XSolidBrush & $-$ & \XSolidBrush   \\
		\hline
		\textbf{Helen}~\cite{le2012interactive}    & $2,330$  & \XSolidBrush& $194$ & \CheckmarkBold &\XSolidBrush & $-$ & \XSolidBrush  \\
		\hline
	   
	   	\textbf{300W}~\cite{sagonas2013300}  & $3,837$  & \XSolidBrush & $68$ & \XSolidBrush & \XSolidBrush & $-$ & \XSolidBrush   \\
	   	\hline 
	   	\textbf{300W-LP}~\cite{zhu2016face}  & $61,225$  & \XSolidBrush & \XSolidBrush & \XSolidBrush & \XSolidBrush & $-$ & \XSolidBrush   \\
	   \hline   
	   \textbf{LS3D-W}~\cite{bulat2017far}  & $230,000$  & \XSolidBrush & $68$ & \XSolidBrush & \XSolidBrush & $-$ & \XSolidBrush   \\
		\hline
		
		\textbf{LS3D-W balanced}~\cite{bulat2017far}  & $7,200$  & \XSolidBrush & $68$ & \XSolidBrush & \XSolidBrush & $-$ & \XSolidBrush   \\
		\hline
		\textbf{CASIA-WebFace}~\cite{yi2014learning}  & $494, 414$  & \XSolidBrush &  2 &  \XSolidBrush & \CheckmarkBold  & $250 \times 250$ & \XSolidBrush  \\ \hline
		\textbf{CelebA}~\cite{liu2015deep} & $202, 599$ & $40$ & $5$ & \XSolidBrush & \CheckmarkBold &$-$ & \XSolidBrush   \\ \hline
		\textbf{IMDB-WIKI}~\cite{rothe2015dex} & $524,230$ & \XSolidBrush & \XSolidBrush &\XSolidBrush &\XSolidBrush & $-$ & \XSolidBrush \\ \hline 
		\textbf{VGGFace}~\cite{parkhi2015deep} & $2,600,000$ & \XSolidBrush & \XSolidBrush& \XSolidBrush &\CheckmarkBold & $-$ &  \XSolidBrush \\ \hline 
		
		\textbf{Menpo}~\cite{zafeiriou2017menpo} &  $8,979$ & \XSolidBrush & $68/39$ & \XSolidBrush & \XSolidBrush  & $-$ & \XSolidBrush   \\ \hline	
		
		\textbf{VGGFace2}~\cite{cao2018vggface2}   &  $3,310,000$  & \XSolidBrush& \XSolidBrush & \XSolidBrush& \CheckmarkBold & $-$ & \XSolidBrush  \\ \hline	
	
\textbf{CelebA-HQ}~\cite{karras2018progressive}  &  $30,000$  & \XSolidBrush& $5$ &\XSolidBrush & \XSolidBrush 
		& $1024 \times 1024$ & \XSolidBrush   \\ \hline
		\textbf{FFHQ}~\cite{karras2019style}  &  $70,000$  & \XSolidBrush& $68$ &\XSolidBrush & \XSolidBrush 
		& $1024 \times 1024$ & \XSolidBrush   \\ \hline
		
		\textbf{EDFace-Celeb-1M}~\cite{zhang2022edface}  &  $1,700,000$  & \XSolidBrush & \XSolidBrush& \XSolidBrush & \XSolidBrush
		& $-$ & \CheckmarkBold  \\ \hline
		
		\textbf{EDFace-Celeb-1M (BFR128)}~\cite{zhang2022blind}  &  $1,505,888$  & \XSolidBrush  
		& \XSolidBrush & \XSolidBrush & \XSolidBrush & $128 \times 128$ & \CheckmarkBold   \\ \hline
		
		\textbf{EDFace-Celeb-150K (BFR512)}~\cite{zhang2022blind}  &  $148,962$  &  \XSolidBrush& \XSolidBrush & \XSolidBrush & \XSolidBrush & $512 \times 512$ & \CheckmarkBold   \\ 

		\hline
	\end{tabular}}
\vspace{-3mm}
\end{table*}
\subsection{Datasets}

To facilitate training and evaluation in face restoration, numerous benchmark datasets have been proposed. Table~\ref{tab:summary_dataset} summarizes the key datasets together with their typical availability for research use and the main annotations or metadata provided, which are detailed below. \textbf{BioID}~\cite{jesorsky2001robust} contains 1,521 grayscale images from 23 subjects and is commonly used for classical face analysis. \textbf{LFW}~\cite{huang2008labeled} includes 13,233 images of 5,749 people and provides identity labels and verification protocols. It is widely available for research use. \textbf{AFLW}~\cite{koestinger2011annotated} offers 25,993 images with face bounding boxes and up to 21 annotated landmarks per image, covering a wide range of poses and expressions. \textbf{Helen}~\cite{le2012interactive} consists of 2,330 high-resolution images with dense landmark annotations, such as 194 landmarks per face. \textbf{300W}~\cite{sagonas2013300} provides 3,837 images with 68-point landmark annotations. Its extended version, \textbf{300W-LP}~\cite{zhu2016face}, contains 61,225 images rendered with varied poses. \textbf{LS3D-W}~\cite{bulat2017far} offers about 230,000 face images with 3D landmark annotations. Its balanced subset contains 7,200 images evenly distributed across pose ranges. \textbf{CASIA-WebFace}~\cite{yi2014learning} comprises 494,414 face images of 10,575 subjects at $250 \times 250$ resolution and provides identity labels. It is commonly used for face recognition pretraining under research access policies.  \textbf{CelebA}~\cite{liu2015deep} contains 202,599 images from 10,177 identities, each annotated with 40 attributes and 5 key points, and is widely available for research use. Based on it, CelebA-Test~\cite{li2020blind,wang2022restoreformer} is synthesized with 3,000 CelebA-HQ images for model evaluation. \textbf{IMDB-WIKI}~\cite{rothe2015dex} provides 524,230 images sourced from IMDB and Wikipedia with metadata such as age labels, and is commonly used for age estimation.  \textbf{VGGFace}~\cite{parkhi2015deep} includes 2.6 million images from 2,622 identities and provides identity labels under dataset usage policies.  \textbf{Menpo}~\cite{zafeiriou2017menpo} has 8,979 images with landmark annotations, such as 68-point or 39-point landmarks depending on visibility. \textbf{VGGFace2}~\cite{cao2018vggface2} contains 3.31 million images of 9,131 subjects with diverse variations and manually validated bounding boxes and identity labels under dataset usage policies.  \textbf{CelebA-HQ}~\cite{karras2018progressive} is a high-resolution version of CelebA at 1024$\times$1024 and is widely used for face generation and restoration.  \textbf{FFHQ}~\cite{karras2019style} consists of 70,000 high-quality images from Flickr with diverse demographics and accessories, and is widely used as a high-quality face prior dataset. \textbf{EDFace-Celeb-1M}~\cite{zhang2022edface} contains 1.7M images with racial diversity, including 1.5M paired low- and high-resolution images and 140K real-world tiny faces for evaluation. \textbf{EDFace-Celeb-1M (BFR128)}~\cite{zhang2022blind} is designed for blind face restoration with synthetic degradations, including blur, noise, low resolution, JPEG artifacts, and combinations, containing 1.5M images at $128\times128$ across multiple tasks. \textbf{EDFace-Celeb-150K (BFR512)}~\cite{zhang2022blind} shares similar degradation settings with BFR128 but features 149K higher-resolution images at $512\times512$, with 132K for training and 17K for testing.

\section{Performance Evaluation}\label{sec:performance_evaluation}

In this section, we present a systematic benchmark evaluation of representative face restoration methods. We first introduce the evaluation settings and datasets, then report quantitative comparisons under synthetic and real-world scenarios, and finally provide qualitative visual results together with efficiency and complexity analysis.

\subsection{Representative Methods}

To better understand the landscape of deep learning-based face restoration, we evaluate a diverse set of recent methods across both synthetic datasets (EDFace-Celeb-1M (BFR128)~\cite{zhang2022blind}, EDFace-Celeb-150K (BFR512)~\cite{zhang2022blind}, CelebA-Test~\cite{liu2015deep}, CelebA-HQ~\cite{karras2018progressive}, FFHQ~\cite{karras2019style}) and real-world datasets (LFW-Test~\cite{huang2008labeled}, CelebChild~\cite{wang2021towards}, WebPhoto~\cite{wang2021towards}). The selected methods span various categories and have publicly available code.

\textbf{GAN-based methods:}  
TDTN~\cite{wan2020bringing} uses a triple domain translation network for old photo restoration.  
DFDNet~\cite{li2020blind} leverages facial dictionaries for reference-based restoration.  
mGANprior~\cite{gu2020image} applies GAN inversion via multi-latent code sampling.  
HiFaceGAN~\cite{yang2020hifacegan} employs hierarchical semantic guidance.  
PULSE~\cite{menon2020pulse} restores via latent space exploration.  
PSFR-GAN~\cite{chen2021progressive}, GPEN~\cite{yang2021gan}, and GFP-GAN~\cite{wang2021towards} are recent GAN-based state-of-the-art frameworks.

\textbf{Transformer-based methods:}  
STUNet~\cite{zhang2022blind} and RestoreFormer~\cite{wang2022restoreformer} introduce Transformer structures with spatial attention for blind face restoration.  
CodeFormer~\cite{zhou2022towards} predicts discrete latent codes for high-fidelity reconstruction.

\textbf{Other learning paradigms:}  
VQFR~\cite{gu2022vqfr} and DAEDR~\cite{tsai2023dual} incorporate vector quantization to guide detail recovery.  
DifFace~\cite{yue2022difface} and OSDFace~\cite{wang2025osdface} introduce a diffusion-based framework for blind face restoration.

\begin{table*}[t]
\centering  
\caption{Performance comparison among representative BFR methods. The training sets are EDFace-Celeb-1M (BFR128) and EDFace-Celeb-150K (BFR512) respectively. The best and the second best performance values are highlighted and underlined respectively. Note that DFDN can only generate $512 \times 512 $ face results for any input image, thus we do not report its results on the EDFace-Celeb-1M (BFR128) dataset.} 
 \vspace{-3mm}
\resizebox{\linewidth}{!}{\begin{tabular}{c|l|ccccc|ccccc} 
\hline
\multirow{2}{*}{Task} & \multirow{2}{*}{Methods} &  \multicolumn{5}{c|}{ EDFace-Celeb-1M (BFR128)} &  \multicolumn{5}{c}{ EDFace-Celeb-150K (BFR512)} \\ \cline{3-7} \cline{8-12}
 &  & PSNR$\uparrow$ & SSIM$\uparrow$ & MS-SSIM$\uparrow$ & LPIPS$\downarrow$  & NIQE$\downarrow$ & PSNR$\uparrow$ & SSIM$\uparrow$ & MS-SSIM$\uparrow$ & LPIPS$\downarrow$  & NIQE$\downarrow$ \\ 

\hline
\multirow{6}{*}{Face Deblurring} 
& DFDNet \cite{li2020blind}  & $-$ & $-$ & $-$ & $-$ & $-$ & 25.4072&0.6512&0.8724 &0.4008 & \textbf{7.8913}\\
& HiFaceGAN \cite{yang2020hifacegan} &22.4598&0.7974 & 0.9420         &0.0739 & 8.7261 & 26.7421 &\textbf{ 0.8095}&\textbf{0.9382} &0.2029&16.6642 \\
& PSFR-GAN \cite{chen2021progressive}  &\textbf{29.1411} &\textbf{0.8563}& \textbf{0.9818}       & \textbf{0.0480} &9.0008 &27.4023  &0.7604 & 0.9155 &0.2292  &17.4076 \\
& GFP-GAN~\cite{wang2021towards}    &25.3822 &0.7461&0.9534    & 0.0704    &12.3608  
& \underline{28.8166} & 0.7709 & 0.9180 &\underline{0.1721} &15.5942 \\

& GPEN~\cite{yang2021gan}   & 24.9091 &0.7307 &  0.9500     &0.0887  &\underline{8.2288}&27.0658 & 0.7175 & 0.8928 & 0.2188 & 15.3187  \\
&RestoreFormer~\cite{wang2022restoreformer}&27.2223 &0.7560 & 0.9587& 0.0692&9.0112 & 28.7231&0.7519 & 0.9121& 0.1825 & 15.3217\\
&DAEFR~\cite{tsai2023dual} &24.9128 & 0.7386& 0.9591& 0.0886& 8.2567& 26.9810&0.7210 & 0.8934&0.1791 &14.9012\\
&OSDFace~\cite{wang2025osdface} &22.4139 & 0.7891& 0.9328& 0.0671&8.2109 & 25.0123& 0.6439&0.8689 & 0.1725&14.7899\\
&VQFR~\cite{gu2022vqfr} &27.3524 & 0.7989&0.9618 & \underline{0.0632}&\textbf{8.1284} & 28.4513& 0.7218& 0.9001& \textbf{0.1698} &15.3019\\
& STUNet~\cite{zhang2022blind}   &\underline{27.3912} & \underline{0.8080} & \underline{0.9669} & 0.2019   &  12.2652 &\textbf{29.5572}&\underline{0.8052} &\underline{0.9289} &0.3381&\underline{14.7874} \\ 
\hline
\hline
\multirow{6}{*}{Face Denoising} 
& DFDNet \cite{li2020blind} & $-$ & $-$ & $-$ & $-$ & $-$ & 24.3618&0.5738 &0.8423&0.3238&\textbf{7.7809}\\

& HiFaceGAN \cite{yang2020hifacegan} &26.2976&0.8801 &0.9663       &0.0306 & \textbf{7.2432} & 30.0409 &\underline{0.8731} & 0.9563 & 0.1439 & 16.7363\\

& PSFR-GAN \cite{chen2021progressive} &\underline{33.1007} &0.8563& 0.9818 & 0.0480 &9.0008  &28.5397 & 0.8232 & 0.9390 & 0.2208 & 19.4719 \\

& GFP-GAN \cite{wang2021towards}   & 31.1053 & 0.8802 & 0.9849 &0.0234    &\underline{7.9522} & \underline{33.2020} & 0.8711 & \underline{0.9582} & \textbf{0.1259} & 15.8440 \\

& GPEN \cite{yang2021gan}  &33.0744 &\underline{0.9086}&\underline{0.9871}     & \underline{0.0211}    &8.0616  &32.3736 &0.8517 & 0.9506 & 0.1555 & 15.6820 \\
&RestoreFormer~\cite{wang2022restoreformer}& 32.9817 &0.8810 &0.9852 &0.0312&8.0412 & 33.1848 & 0.8698& 0.9517&\underline{0.1369} & \underline{15.1598}\\
&DAEFR~\cite{tsai2023dual}  &30.4831 & 0.8412& 0.9452&0.0238 & 7.9936&31.2813 & 0.8423&0.9481 & 0.1390&15.6034\\
&OSDFace~\cite{wang2025osdface} &26.1728 & 0.8329& 0.9140&0.0221 &7.9923& 24.5612&0.5819 &0.8517 &0.1388 & 15.1621\\\
&VQFR~\cite{gu2022vqfr} &33.0114 &0.9001 & 0.9861& \textbf{0.0208}& 7.9812&33.1247 &0.8685 &0.9498 & 0.1401 & 16.0212\\
& STUNet~\cite{zhang2022blind}   & \textbf{34.8914} &\textbf{0.9302} & \textbf{0.9900 }   &   0.0331 & 8.5349  & \textbf{34.5500} & \textbf{0.8848} & \textbf{0.9587} & 0.1787 & 16.5480      \\ 
\hline
\hline
\multirow{6}{*}{Face Artifact Removal}
& DFDNet \cite{li2020blind}  & $-$ & $-$ & $-$ & $-$ & $-$ &27.4781&0.7845&0.9409 &0.2241&\textbf{7.5553}\\
& HiFaceGAN \cite{yang2020hifacegan}    &23.8228 &0.8531 & 0.9567    &0.0453 & \textbf{7.6479}
&27.1164 & 0.8897&0.9635 &0.1241&18.7117  \\
& PSFR-GAN \cite{chen2021progressive} &\underline{31.9455} &\underline{0.8899}& \underline{0.9887}   & \textbf{0.0190} &8.3158 &29.4285 &0.9101 &0.9719 &0.1245 &\underline{15.9760} \\
& GFP-GAN \cite{wang2021towards}   &31.0910 &0.8804&0.9874& \underline{0.0227}   &\underline{7.8027 } &\underline{35.7201} &\underline{0.9144} & \underline{0.9780}&\textbf{ 0.0842} &16.8320   \\
& GPEN \cite{yang2021gan}  & 30.5753 &0.8556 &0.9837         &0.0241  &7.8074 &33.8355 &0.8701 &0.9657  &\underline{0.0986} &16.9854  \\
&RestoreFormer~\cite{wang2022restoreformer}&30.9811 & 0.8798 & 0.9851&0.0231 & 7.8036 & 35.3147 & 0.9098& 0.9710&0.1023 & 16.9134\\
&DAEFR~\cite{tsai2023dual}  &29.1203 &0.8649 & 0.9840& 0.0245&7.8032 & 29.5210&0.8713 & 0.9612& 0.0993&15.9012\\
&OSDFace~\cite{wang2025osdface} &22.4312 &0.8313 &0.9312 &0.0230 &7.8028 &26.1453 & 0.6910& 0.8170& 0.0988&15.8018\\
&VQFR~\cite{gu2022vqfr} &30.9711 & 0.8791& 0.9843& 0.0238&  7.8051& 35.2956 & 0.9038&0.9701 &0.1192 & 16.9321\\
& STUNet~\cite{zhang2022blind}   &\textbf{33.2082} & \textbf{0.9171} & \textbf{0.9912}     &  0.0582   & 10.5596       &\textbf{36.5017} &\textbf{ 0.9246} & \textbf{0.9799} &0.1411 &16.0487 \\  
\hline
\hline
\multirow{6}{*}{Face Super-Resolution}
& DFDNet \cite{li2020blind}  & $-$ & $-$ & $-$ & $-$ & $-$ &26.8691&0.7405&0.9224& 0.2620& \textbf{7.4796}\\
& HiFaceGAN \cite{yang2020hifacegan}&24.2965&\underline{0.7792} & 0.9493         &0.0911 & 8.4801 &26.6103 & 0.8480 &0.9476 &0.1681 &15.8911 \\
& PSFR-GAN \cite{chen2021progressive} &23.9671 &0.6858& 0.9381      & 0.1364 &\textbf{7.4807}  &33.1233  &0.8588 &0.9602  &\underline{0.1331} &16.7143 \\
& GFP-GAN \cite{wang2021towards}   &\underline{25.7118} &0.7558&0.9492     & \textbf{   0.0762} &11.4428  &\underline{33.4217}  &\underline{0.8629}&\underline{0.9610} &\textbf{0.1127} &16.8970       \\
& GPEN \cite{yang2021gan}   &25.0208 &0.7306&0.9448     & \underline{0.0843}   &\underline{7.9052}  &31.3507&0.8273 &0.9501 &0.1357&\underline{15.7813}  \\
&RestoreFormer~\cite{wang2022restoreformer}& 25.0609& 0.7398& \underline{0.9497} &0.0878& 7.9671& 33.1213&0.8580 &0.9596 & 0.1353 & 15.8012\\
&DAEFR~\cite{tsai2023dual}  & 24.9817& 0.7512& 0.9461&0.0860 &7.9123 &26.7133 & 0.8610& 0.9614& 0.1378&15.8913\\\
&OSDFace~\cite{wang2025osdface} &23.8716& 0.7210&0.9011 & 0.0844& 7.9027& 26.1267& 0.7015& 0.9120& 0.1335&15.7932\\
&VQFR~\cite{gu2022vqfr} &25.0456 & 0.7387& 0.9490& 0.0897&7.9682 & 33.1028& 0.8573&0.9590 &0.1371 &15.8023\\
& STUNet~\cite{zhang2022blind}   &\textbf{27.1206} & \textbf{0.8037} & \textbf{0.9566}     &   0.2018  & 12.7177  &\textbf{33.9060} &\textbf{0.8809} &\textbf{0.9636} &0.2235 &17.0899 \\ 
\hline
\hline
\multirow{6}{*}{Blind Face Restoration}
& DFDNet \cite{li2020blind}  & $-$ & $-$ & $-$ & $-$ & $-$ &23.9349&0.5573&0.8053& 0.4231&\textbf{9.0084}\\
& HiFaceGAN \cite{yang2020hifacegan}&22.2179&\textbf{0.7088} & 0.9128         &0.1528 & 9.6864 &25.3083 & 0.7260&\underline{0.8701} &\underline{0.3012} &14.7883 \\
& PSFR-GAN \cite{chen2021progressive} &22.2620 &0.5199& 0.8811      & 0.3558 &\underline{8.3706}  &26.2998 &0.6934 & 0.8581&0.3167 &17.1906  \\
& GFP-GAN \cite{wang2021towards}  &\underline{23.4159} &0.6707&\underline{0.9185}  &\textbf{0.1354} &12.6364  &\textbf{28.4809}  & \textbf{0.7857}&\textbf{0.9255}  &\textbf{0.2171} &\underline{14.4933}  \\
& GPEN \cite{yang2021gan}  & 22.9731 &0.6348 &0.9119 &\underline{0.1387}   &\textbf{8.0709} &25.5778 &0.6721 &0.8448 & 0.3113 & 15.8422      \\
&RestoreFormer~\cite{wang2022restoreformer}& 23.4017& 0.6891&0.9169 & 0.1360& 8.4213& 27.0114& 0.7332& 0.8632& 0.4187&14.9811 \\
&DAEFR~\cite{tsai2023dual}  &22.0126 &0.6123 &0.9013 &0.1399 & 8.4012&25.2912 &0.7234 &0.8640 & 0.3459&14.6891\\
&OSDFace~\cite{wang2025osdface} & 21.1256& 0.5816& 0.8910& 0.1390&8.3788 &23.1234 &0.5613 &0.8077 &0.3015 &14.5016\\
&VQFR~\cite{gu2022vqfr} & 23.4001& 0.6884& 0.9168& 0.1365& 8.4229&27.0107 &0.7308 & 0.8629 &0.4205 &14.9895\\
& STUNet~\cite{zhang2022blind}  &\textbf{24.5500} & \underline{0.6978} & \textbf{0.9225}&  0.3523   & 13.0601  &\underline{27.1833} &\underline{0.7346} &0.8654 &0.4457& 17.0305 \\ 
\hline
\end{tabular}}
\label{table:128_1}
\vspace{-4mm}
\end{table*}

\subsection{Experimental Setting and Metric} 
\textbf{Setting}. To provide a clear view of existing face restoration methods, we use both synthetic datasets (EDFace-Celeb-1M (BFR128)~\cite{zhang2022blind}, EDFace-Celeb-150K (BFR512)~\cite{zhang2022blind}, FFHQ~\cite{karras2019style}, CelebA-HQ~\cite{karras2018progressive}, and CelebA-Test~\cite{liu2015deep}) and real-world datasets (LFW-Test, CelebChild-Test, and WebPhoto-Test~\cite{wang2021towards}) for training and evaluation. Specifically, we conduct two distinct experimental settings for the purpose of model training and testing. The first scheme is based on EDFace benchmarks~\cite{zhang2022blind} and is used for Table~\ref{table:128_1} to evaluate multiple face restoration task settings under controlled synthetic degradations with paired ground truth. The second scheme is based on the FFHQ benchmark used in representative works~\cite{li2020blind,wang2021towards,wang2022restoreformer} and is used for Table~\ref{tab:merged} to evaluate both a synthetic test set and real-world test sets for generalization. The details of these settings are outlined below.


Experimental Scheme One. We use EDFace-Celeb-1M (BFR128) and EDFace-Celeb-150K (BFR512) with their official training and testing splits. The paired targets enable reporting full reference metrics for face deblurring, face denoising, face artifact removal, face super-resolution, and blind face restoration. The corresponding results are reported in Table~\ref{table:128_1}. Specifically, we adopt the code and pre-trained models provided in \cite{zhang2022blind}. For a fair comparison, in this scheme, the compared methods are trained on the official training split and evaluated on the official test split for each task setting. When official models or configurations are released, we follow the official implementation and training settings.

Experimental Scheme Two: In~\cite{li2020blind,wang2021towards,wang2022restoreformer}, they first synthesize degraded face images on the FFHQ~\cite{karras2019style} dataset via the degradation model. In the degradation model, the parameters $\sigma$, $\delta$, $r$ and $q$ are randomly sampled from $\{0.2: 10\}$, $\{1:8\}$, $\{0:20\}$, and $\{60:100\}$, respectively. Then, they use paired face images to train networks. In addition, one synthetic dataset (CelebA-Test~\cite{liu2015deep}) and three real-world datasets (LFW-Test, CelebChild-Test, and WebPhoto-Test~\cite{wang2021towards}) are used as testing datasets to measure the performance of models. Note that these testing sets have no overlap with the FFHQ dataset. In this scheme, we use official released checkpoints when they are available. If an official checkpoint is not provided, we train the model using the official implementation under the same FFHQ based setting. We do not fine-tune on CelebA-Test or the real-world test sets. The corresponding results are reported in Table~\ref{tab:merged}.

\textbf{Metric}. The full-reference and non-reference metrics are employed to benchmark these methods in the experiments. The full-reference metrics contain PSNR, SSIM, MS-SSIM, and LPIPS. These full-reference metrics measure the visual quality in different aspects, including pixels, structure, and human perception. The non-reference metrics consist of NIQE and FID, which can be used for real-world datasets without ground truth.

\begin{table*}[t]
  \centering
  \caption{Quantitative comparisons on synthetic (CelebA-Test) and real-world datasets (LFW-Test, CelebChild-Test, WebPhoto-Test). Metrics include FID, PSNR, SSIM, LPIPS, and NIQE. The red and blue colors indicate the best and second-best performance, respectively. All models are trained or finetuned on face images synthesized from FFHQ~\cite{karras2019style}.}
  \vspace{-3mm}
  \resizebox{\linewidth}{!}{
  \begin{tabular}{c|cccc|cc|cc|cc}
    \hline
    \multirow{3}{*}{\textbf{Methods}} 
    & \multicolumn{4}{c|}{\textbf{CelebA-Test (Synthetic)}} 
    & \multicolumn{2}{c|}{\textbf{LFW-Test}} 
    & \multicolumn{2}{c|}{\textbf{CelebChild-Test}} 
    & \multicolumn{2}{c}{\textbf{WebPhoto-Test}} \\ 
    \cline{2-11}
    & FID$\downarrow$ & PSNR$\uparrow$ & SSIM$\uparrow$ & LPIPS$\downarrow$
    & FID$\downarrow$ & NIQE$\downarrow$ 
    & FID$\downarrow$ & NIQE$\downarrow$ 
    & FID$\downarrow$ & NIQE$\downarrow$ \\
    \hline
    Input 
    & 132.69 & 24.96 & 0.6624 & 0.4989 
    & 137.56 & 11.214 & 144.42 & 9.170 & 170.11 & 12.755 \\
    \hline
    TDTN~\cite{wan2020bringing}           
    & 70.21 & 23.00 & 0.6189 & 0.4778 
    & 73.19 & 5.034 & 115.70 & 4.849 & 100.40 & 5.705 \\
    PULSE~\cite{menon2020pulse}                
    & 67.75 & 21.61 & 0.6287 & 0.4657 
    & 64.86 & 5.097 & 102.74 & 5.225 & 86.45 & 5.146 \\
    mGANprior~\cite{gu2020image}               
    & 82.27 & 24.30 & \textcolor{blue}{0.6758} & 0.4584 
    & 73.00 & 6.051 & 126.54 & 6.841 & 120.75 & 7.226 \\
    DFDNet~\cite{li2020blind}                  
    & 52.92 & 24.10 & 0.6092 & 0.4478 
    & 62.57 & 4.026 & 111.55 & 4.414 & 100.68 & 5.293 \\
    HiFaceGAN~\cite{yang2020hifacegan}         
    & 66.09 & \textcolor{red}{24.92} & 0.6195 & 0.4770 
    & 64.50 & 4.510 & 113.00 & 4.855 & 116.12 & 4.885 \\
    PSFR-GAN~\cite{chen2021progressive}        
    & 43.88 & 24.45 & 0.6308 & 0.4186 
    & 51.89 & 5.096 & 107.40 & 4.804 & 88.45 & 5.582 \\
    GFP-GAN~\cite{wang2021towards}             
    & 42.39 & \textcolor{blue}{24.46} & 0.6684 & 0.3551 
    & 49.96 & 3.882 & 111.78 & \textcolor{blue}{4.349} & 87.35 & 4.144 \\
    RestoreFormer~\cite{wang2022restoreformer} 
    & 41.45 & 24.42 & 0.6404 & 0.3650 
    & 47.75 & 4.168 & \textcolor{blue}{101.22} & 4.582 & 77.33 & 4.459 \\
    CodeFormer~\cite{zhou2022towards}          
    & 60.62 & 22.18 & 0.6104 & \textcolor{red}{0.2993} 
    & 53.83 & 4.473 & 119.13 & 4.905 & 86.10 & 4.628 \\

     DAEFR~\cite{tsai2023dual} & 52.06 & 19.92 &0.5531 & 0.3884&47.53& \textcolor{red}{3.479} & 103.54& 4.632 &\textcolor{blue}{75.47}&\textcolor{blue}{3.933}\\ 

    OSDFace~\cite{wang2025osdface} & 45.41 & 19.61 & 0.5498& \textcolor{blue}{0.3365} &\textcolor{red}{44.63} &3.871 &102.28 &4.546 &84.60 &3.986 \\ 
    
    VQFR~\cite{gu2022vqfr}                     
    & \textcolor{blue}{41.28} & 24.14 & 0.6360 & 0.3515 
    & 50.64 & \textcolor{blue}{3.589} & 105.18 & \textcolor{red}{3.936} & \textcolor{red}{75.38} & \textcolor{red}{3.607} \\
    DifFace~\cite{yue2022difface}              
    & \textcolor{red}{18.27} & 24.08 & \textcolor{red}{0.7036} & 0.4354 
    & \textcolor{blue}{45.23} & 3.834 & \textcolor{red}{96.47} & 4.394 & 85.52 & 4.043 \\
    \hline
  \end{tabular}
  }
  \label{tab:merged}
\end{table*}

\subsection{Quantitative Evaluation}
We evaluate several state-of-the-art face restoration methods on both the synthetic and real-world datasets quantitatively regarding tasks including face deblurring, face denoising, face artifact removal, face super-resolution, and blind face restoration.

Table~\ref{table:128_1} reports the quantitative results of eight face restoration methods on the EDFace-Celeb-1M and EDFace-Celeb-150K datasets, where face deblurring, face denoising, face artifact removal, face super-resolution, and blind face restoration refer to five face restoration tasks related to the degradation from blur, noise, JPEG, low resolution, and a mix of them, respectively. For the comparison results in terms of PSNR, SSIM, MS-SSIM, LPIPS, and NIQE, we have the following findings. (i) In terms of PSNR, SSIM, and MS-SSIM, the Transformer-based method STUNet~\cite{zhang2022blind} is very competitive and outperforms the best face prior based methods. Specifically, STUNet~\cite{zhang2022blind} achieves the best
performance on face denoising, face artifact removal, and face super-resolution tasks, and it also achieves the second best performance in face deblurring and blind face restoration tasks. Compared with other face restoration methods (DFDNet~\cite{li2020blind}, HiFaceGAN~\cite{yang2020hifacegan}, PSFR-GAN~\cite{chen2021progressive}, GFP-GAN~\cite{wang2021towards}, GPEN~\cite{yang2021gan}, RestorFormer~\cite{wang2022restoreformer}, DAEFR~\cite{tsai2023dual}, OSDFace~\cite{wang2025osdface}, and VQFR~\cite{gu2022vqfr}), though STUNet~\cite{zhang2022blind} does not explicitly consider the face-related prior, it still achieves outstanding performance in many face restoration tasks. It demonstrates that it is very important to choose a reasonable network architecture, and a well-designed architecture will easily result in stronger performance. This observation will inspire us to design deep networks based on a strong backbone network.
(ii) In terms of LPIPS and NIQE (non-reference quantitative metrics), GFP-GAN~\cite{wang2021towards} and GPEN~\cite{yang2021gan} achieve the best or second best performance for most face tasks, and HiFaceGAN~\cite{yang2020hifacegan}, PSFR-GAN~\cite{chen2021progressive}, RestorFormer~\cite{wang2022restoreformer}, and VQFR~\cite{gu2022vqfr} place the best or second best on some face tasks. Compared with STUNet, GAN-based methods (HiFaceGAN~\cite{yang2020hifacegan}, PSFR-GAN~\cite{chen2021progressive}, GFP-GAN~\cite{wang2021towards}, and GPEN~\cite{yang2021gan}) obtain better performance. Because GAN based methods are good at generating content pleasing the human visual perception system, easily achieving better performance on non-reference quantitative metrics. This suggests that we should consider more different metrics when carrying out the performance evaluation.

\begin{figure*}[t] 
\begin{center}
 \begin{overpic}[width=0.8\textwidth]{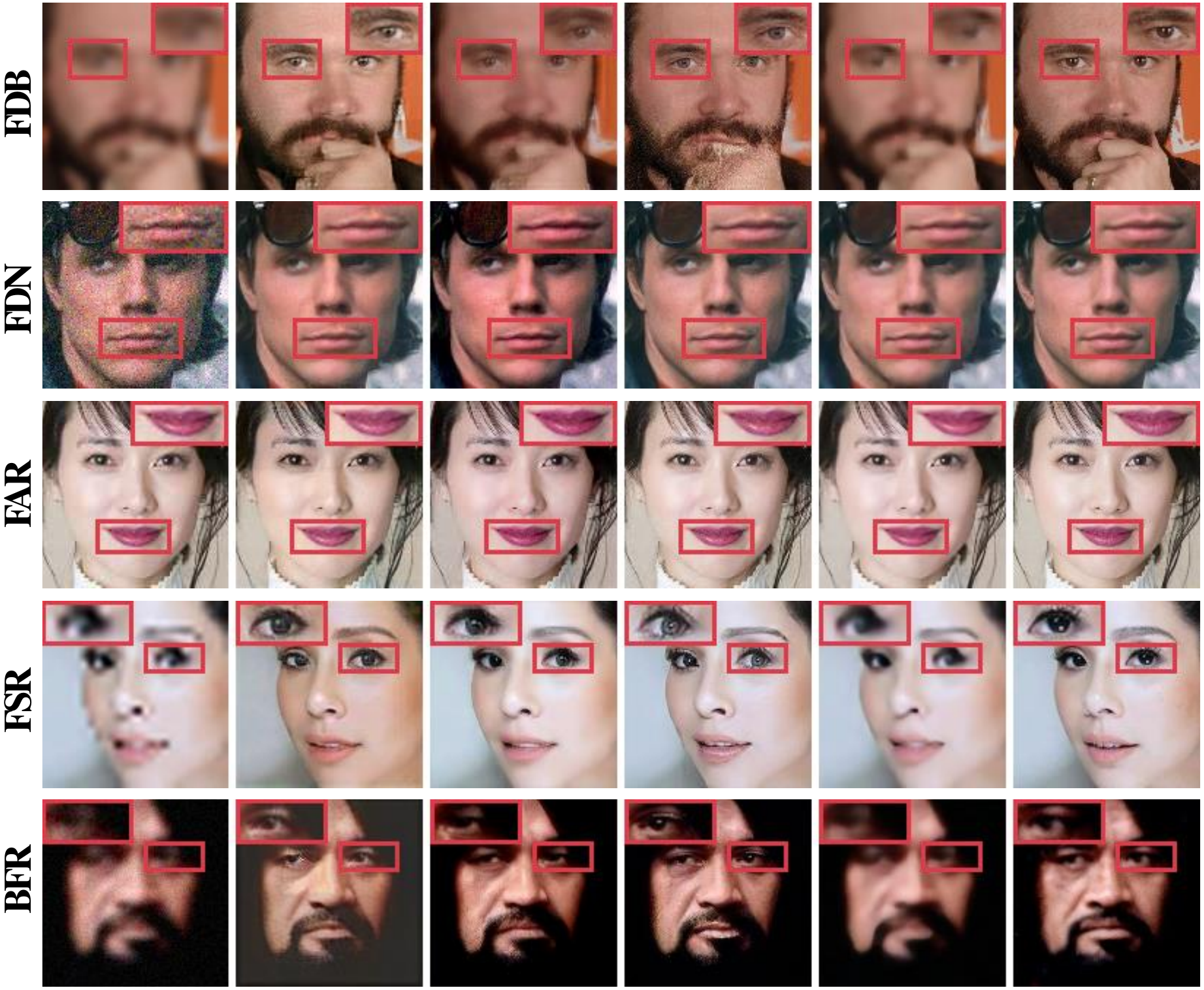}  
 \end{overpic}
 	\caption{Visual comparison on the \textbf{EDFace-Celeb-1M (BFR 128)} dataset. FDB, FDN, FAR, FSR, and BFR indicate face deblurring, face denoising, face artifact removal, face super-resolution, and face restoration, respectively. From left to right are the input, the results of HiFaceGAN, PSFR-GAN, GPEN, and STUNet, and HQ images. }
	\label{fig:128_512_results_a}
\end{center}
\end{figure*}

The quantitative results of the CelebA-Test~\cite{liu2015deep} dataset are shown in Table~\ref{tab:merged}, which includes a comprehensive evaluation of various state-of-the-art methods for blind face restoration. The comparison methods are TDTN~\cite{wan2020bringing}, PULSE~\cite{menon2020pulse}, mGANprior~\cite{gu2020image}, DFDNet~\cite{li2020blind}, HiFaceGAN~\cite{yang2020hifacegan}, PSFR-GAN~\cite{chen2021progressive}, GFP-GAN~\cite{wang2021towards}, RestoreFormer~\cite{wang2022restoreformer}, CodeFormer~\cite{zhou2022towards}, DAEFR~\cite{tsai2023dual}, VQFR~\cite{gu2022vqfr}, and the recent diffusion-based methods DifFace~\cite{yue2022difface} and OSDFace~\cite{wang2025osdface}. Among these methods, DifFace~\cite{yue2022difface}, OSDFace~\cite{wang2025osdface}, and VQFR~\cite{gu2022vqfr} demonstrate competitive performance on the CelebA-Test dataset. Notably, DifFace~\cite{yue2022difface}, as a diffusion-based method, achieves the highest performance in terms of FID and SSIM, signifying its superior ability in blind face restoration. OSDFace~\cite{wang2025osdface} and VQFR~\cite{gu2022vqfr}, on the other hand, secure the second best performance in FID and LPIPS. Regarding PSNR, HiFaceGAN~\cite{yang2020hifacegan} yields the highest performance, while GFP-GAN attains the second best performance. From these results, we can observe that GAN-based methods often obtain better perceptual scores and produce sharper textures due to adversarial learning, but may suffer from potential identity drift or hallucinated details. In contrast, diffusion-based methods (DifFace~\cite{yue2022difface} and OSDFace~\cite{wang2025osdface}) can yield competitive perceptual quality and improved distribution alignment, but typically require iterative sampling at inference, leading to higher computational cost.


To evaluate the effectiveness of existing face restoration methods in real-world scenarios, we systematically assess their performance on three diverse real-world datasets: LFW-Test, CelebChild-Test, and WebPhoto-Test~\cite{wang2021towards}. Table~\ref{tab:merged} reports the quantitative results in terms of FID and NIQE. DAEFR~\cite{tsai2023dual}, OSDFace~\cite{wang2025osdface}, VQFR~\cite{gu2022vqfr}, DifFace~\cite{yue2022difface}, RestoreFormer~\cite{wang2022restoreformer}, and GFP-GAN~\cite{wang2021towards} exhibit competitive performance in real-world face restoration. Specifically, VQFR~\cite{gu2022vqfr} achieves the highest performance on the WebPhoto-Test dataset and obtains the lowest NIQE scores across the LFW-Test and CelebChild-Test datasets. The superior performance of VQFR~\cite{gu2022vqfr} in real-world face image restoration can be attributed to its effective utilization of the vector quantization technique. Furthermore, the recent diffusion-based methods, OSDFace~\cite{wang2025osdface} and DifFace~\cite{yue2022difface}, also demonstrates competitive performance in real-world scenes.  OSDFace~\cite{wang2025osdface} obtains the best FID scores on the LFW-Test dataset, and DifFace~\cite{yue2022difface} attains the best FID scores on the CelebChild-Test dataset. These real-world results further highlight a representative trade-off between GAN-based and diffusion-based methods: GAN-based methods (\eg GFP-GAN~\cite{wang2021towards}) often generate sharper textures and achieve favorable perceptual quality, but may introduce non-authentic details under severe or unseen degradations, whereas diffusion-based methods (\eg OSDFace~\cite{wang2025osdface} and DifFace~\cite{yue2022difface}) can show stronger distribution alignment (see FID metric) and robustness via iterative refinement, at the cost of higher inference latency. Overall, the evaluation on real-world datasets reveals that the vector quantization technique and recent diffusion models present promising directions for addressing real-world face restoration challenges. Additionally, the experiments demonstrate that FID and NIQE results do not always align. Hence, it is crucial to further explore face restoration performance evaluation metrics for future research.


\begin{figure*}[t]  
\begin{center}
 \begin{overpic}[width=0.9\textwidth]{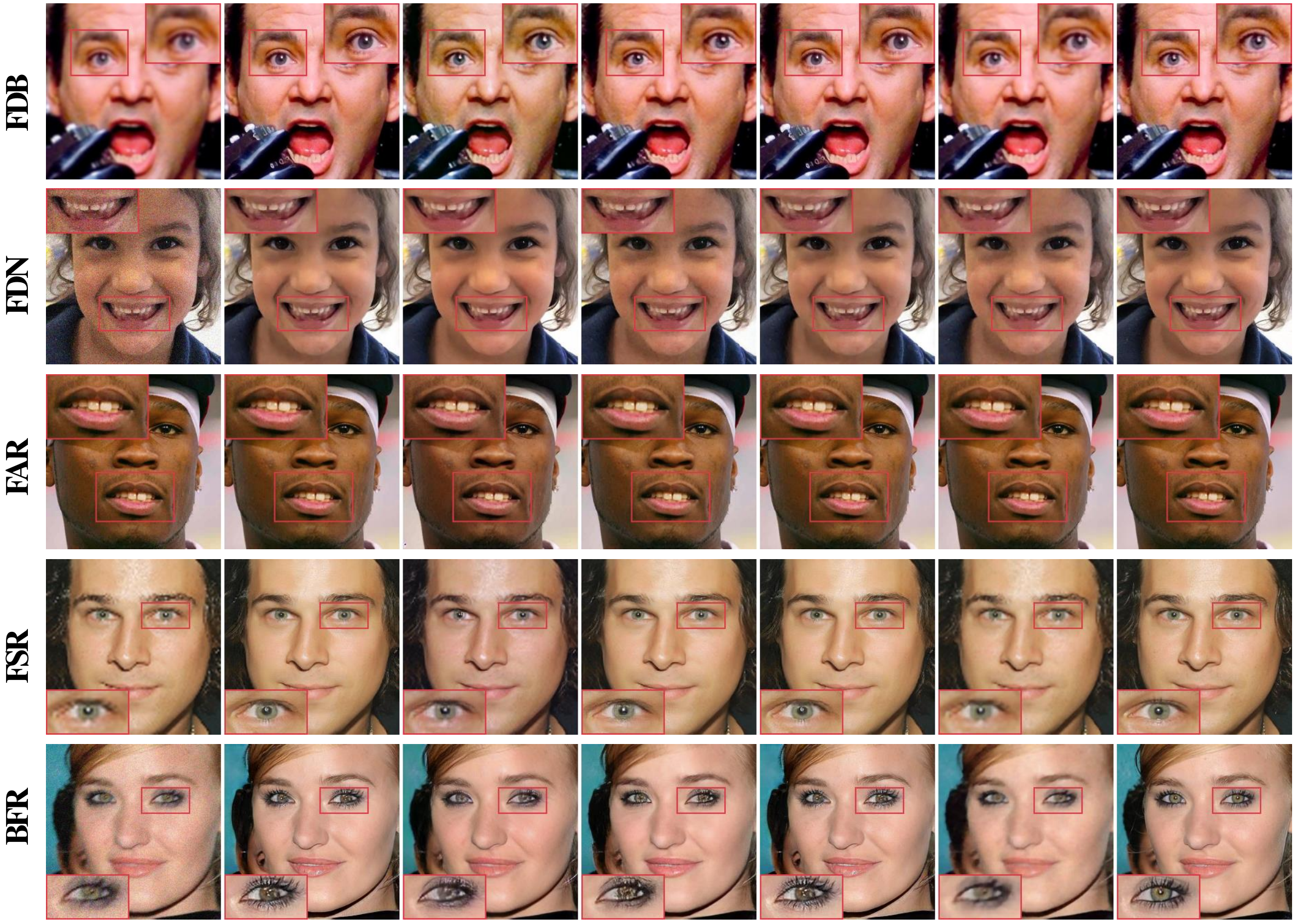}   \end{overpic}
 	\caption{Visual comparison on the \textbf{EDFace-Celeb-150K (BFR512)} dataset. FDB, FDN, FAR, FSR, and BFR indicate face deblurring, face denoising, face artifact removal, face super-resolution, and blind face restoration, respectively. From left to right are the input, the results of DFDNet, HiFaceGAN, PSFR-GAN, GPEN, and STUNet, and HQ images. }
	\label{fig:128_512_results_b}
\end{center}
\end{figure*}

\begin{figure*}[t]
    \centering
    \includegraphics[width=\linewidth]{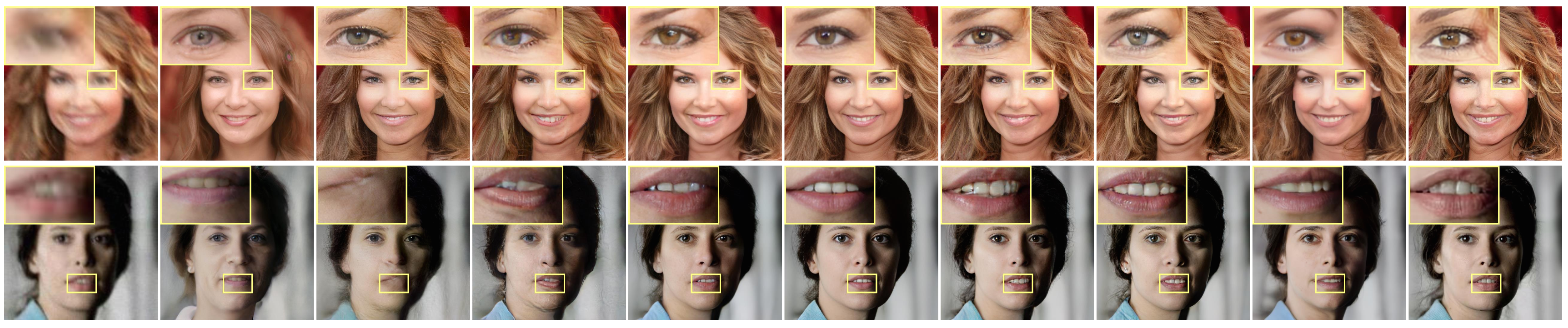}
    \caption{Visual comparison on \textbf{CelebA-Test}. From left to right are the input, the results of PULSE, DFDNet, PSFR-GAN, GFP-GAN, CodeFormer, RestoreFormer, VQFR, DifFace, and HQ images.}
    \label{fig:celeba}
\end{figure*}%

\subsection{Qualitative Evaluation}
Fig.~\ref{fig:128_512_results_a} and Fig.~\ref{fig:128_512_results_b} show some visual results of face deblurring, face denoising, face artifact removal, face super-resolution, and blind face restoration on EDFace-Celeb-1M and EDFace-Celeb-150K datasets, respectively. We can see that face images generated by the GAN-based methods (DFDNet~\cite{li2020blind}, HiFaceGAN~\cite{yang2020hifacegan}, PSFR-GAN~\cite{chen2021progressive}, and GPEN~\cite{yang2021gan}) are more visually pleasing by human visual perception.
For example, for face deblurring in the EDFace-Celeb-1M dataset, HiFaceGAN~\cite{yang2020hifacegan} and GPEN~\cite{yang2021gan} can effectively remove the blur in face images, STUNet~\cite{zhang2022blind} cannot deal with the blur well (see the eyes in the first row of Fig.~\ref{fig:128_512_results_a}). For the most challenging task of blind face restoration, we find that GPEN generates visually more pleasing face images (see the woman's eyes in the last row of Fig.~\ref{fig:128_512_results_b}). The visual results are more consistent with the results of the non-reference metrics. Therefore, non-reference indicators (\eg NIQE, FID) should be fully considered in the performance evaluation.

\begin{figure*}[t]
\begin{center}
 \begin{overpic}[width=\textwidth]{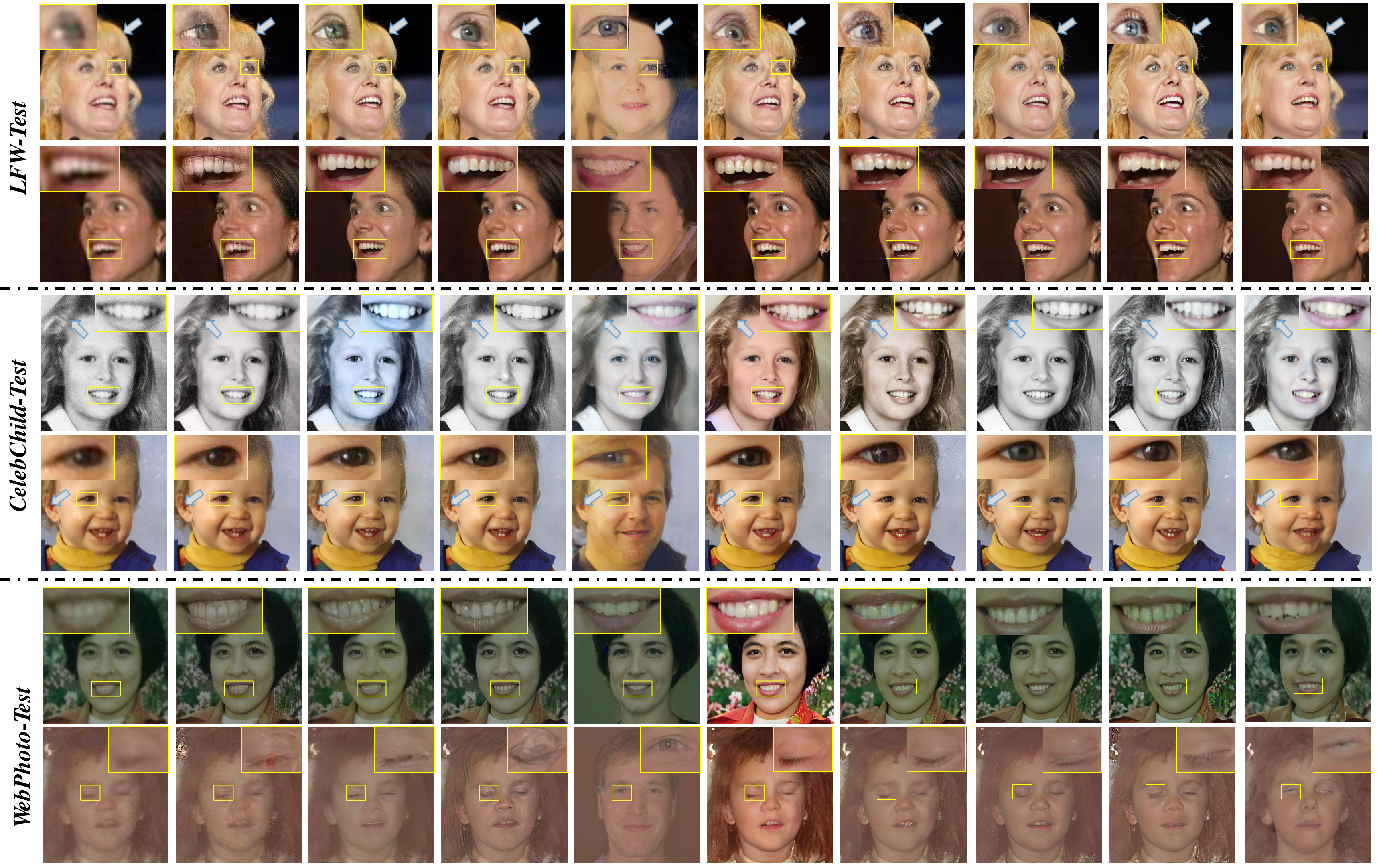}
 \end{overpic}
 	\caption{Visual comparison on \textbf{real-world} \textbf{LFW-Test}, \textbf{CelebChild-Test}, \textbf{WebPhoto-Test} datasets. From left to right are the input, the results of HiFaceGAN, DFDNet, PSFR-GAN, PULSE, GFP-GAN, RestoreFormer, CodeFormer, VQFR, and DifFace.}
	\label{fig:real_results}
\end{center}
\end{figure*}

Fig.~\ref{fig:celeba} presents the visual results on the set of CelebA-Test. In the figure, PULSE can recover face images well. However, it changes the human identity compared with GFP-GAN and RestoreFormer, which indicates that strong generative priors may introduce identity drift in blind restoration. DFDNet~\cite{li2020blind} and PSFR-GAN~\cite{chen2021progressive} cannot recover details of faces well (see the left eye in the first row and the mouth marked yellow box in the second row). The recent state-of-the-art methods RestoreFormer~\cite{wang2022restoreformer}, VQFR~\cite{gu2022vqfr}, and DifFace~\cite{yue2022difface} can generate plausible face images. Among them, GAN-based methods (\eg PULSE~\cite{menon2020pulse}, PSFR-GAN~\cite{chen2021progressive}, GFP-GAN~\cite{wang2021towards}) usually produce sharper textures with feed-forward inference, while the diffusion-based method DifFace refines results progressively and tends to generate more natural details, but typically requires longer inference time due to iterative sampling.

We also evaluate the generalization ability of several representative face restoration methods on real-world datasets: LFW-Test, CelebChild-Test, and WebPhoto-Test. The qualitative comparisons are presented in Fig.~\ref{fig:real_results}. Among the evaluated methods, GFP-GAN~\cite{wang2021towards}, RestoreFormer~\cite{wang2022restoreformer}, and DifFace~\cite{yue2022difface} demonstrate the capability to generate realistic faces in complex real-world scenes. For instance, GFP-GAN~\cite{wang2021towards} effectively enhances color and restores sharp details, such as the woman's mouth in the fifth row and the child's eye in the fourth row, in comparison to all other methods. While the recent diffusion-based method DifFace~\cite{yue2022difface} successfully recovers face images, it struggles to control the facial attributes in the resulting images, as evidenced by the issue with the girl's eye in the last row of comparisons. Overall, GFP-GAN~\cite{wang2021towards} with generative face priors demonstrates superior generalization ability in real-world scenarios. Recent Transformer- and diffusion-based methods also show promising potential for real-world face restoration, benefiting from stronger representation ability and generative priors, but they may still suffer from attribute/identity control issues under severe degradations and typically incur higher computational cost (especially for diffusion-based inference). On the other hand, there is still a need to develop new techniques and models to effectively address the challenges in real-world face restoration.

\begin{table}[t]
  \small 
  \centering 
    \caption{Running time and overhead comparison of typical face restoration methods. The number of parameters (Param) and multiply-accumulate operations (MACs) are used to compute the overhead. MACs are measured on $512 \times 512$ images. We test models with a PC using an NVIDIA GeForce 3090 GPU for fair comparisons.}
    \vspace{-3mm}
    \resizebox{0.8\linewidth}{!}{
    \begin{tabular}{c|c|c|c|c|c|c}
        \hline
        Method & DFDNet~\cite{li2020blind} &  HiFaceGAN~\cite{yang2020hifacegan} & PSFR-GAN~\cite{chen2021progressive} & VQFR~\cite{gu2022vqfr} & TDTN~\cite{wan2020bringing} &GFP-GAN~\cite{wang2021towards}\\
        \hline 
        \hline
        Speed (sec.) & 0.66 & 0.16 & 0.04    &  0.17   & 0.15  &0.06\\
        Params (M) & 133.34 & 130.54 & 45.69 & 71.83   & 97.51 &60.76 \\
        MACs (G) & 608.74 & 697.70 & 102.80   & 1067.18 & 767.04  &85.04\\
        \hline
        \hline
        GPEN~\cite{yang2021gan}  & STUNet~\cite{zhang2022blind} &  RestoreFormer~\cite{wang2022restoreformer} & CodeFormer~\cite{zhou2022towards} & DAEFR~\cite{tsai2023dual}& DifFace~\cite{yue2022difface} &OSDFace~\cite{wang2025osdface}\\
        \hline 
         0.04  & 0.17   & 0.06   & 0.04  & 0.13& 11.38  &0.71\\
         26.23 & 24.81  & 72.37  & 73.57 &112.22 & 159.70 &865.79\\
        17.78 & 334.25 & 340.80 & 292.35 & 449.47& 185.95 &339.24\\        \hline
    \end{tabular}}
    \label{table:speed}
\end{table}%
\subsection{Computational Complexity}

To provide a more comprehensive analysis of the time and complexity of existing face restoration methods, we choose representative FR methods for comparison. Specifically, we select representative methods that cover different model paradigms (CNN, GAN, Transformer, and diffusion model) and have publicly available implementations or reproducible configurations for a fair comparison. These methods include DFDNet~\cite{li2020blind}, HiFaceGAN~\cite{yang2020hifacegan}, PSFR-GAN~\cite{chen2021progressive}, VQFR~\cite{gu2022vqfr}, TDTN~\cite{wan2020bringing}, GFP-GAN~\cite{wang2021towards}, GPEN~\cite{yang2021gan}, STUNet~\cite{zhang2022blind}, RestoreFormer~\cite{wang2022restoreformer}, CodeFormer~\cite{zhou2022towards}, DAEFR~\cite{tsai2023dual}, DifFace~\cite{yue2022difface}, and OSDFace~\cite{wang2025osdface}. Table~\ref{table:speed} shows the running time and overhead of the existing state-of-the-art methods (measured on $512\times512$ images using an NVIDIA GeForce 3090 GPU). Our analysis of the results reveals that the inference time of most methods ranges from 0.04s to 0.66s, where PSFR-GAN, GPEN, and CodeFormer are the fastest (0.04s), while DFDNet is relatively slower (0.66s). In terms of model size, lightweight models such as STUNet (24.81M) and GPEN (26.23M) employ fewer parameters, while DifFace (159.70M), DFDNet (133.34M), and HiFaceGAN (130.54M) have larger parameter sizes. Regarding computational cost (MACs), GPEN requires the fewest MACs (17.78G), and GFP-GAN and PSFR-GAN also have relatively low overhead (85.04G and 102.80G, respectively), whereas VQFR has the highest MACs (1067.18G), followed by TDTN (767.04G), HiFaceGAN (697.70G), and DFDNet (608.74G). However, the existing diffusion-based method DifFace has a much longer inference duration (11.38s) compared to the end-to-end face restoration networks, although its MACs (185.95G) are not the largest, which is mainly due to the iterative denoising/sampling procedure. In comparison, OSDFace, a one-step diffusion model, achieves significantly faster inference (0.13s) while maintaining competitive restoration quality. Some Transformer-based methods, such as STUNet, RestoreFormer, and CodeFormer, also show relatively high computational complexity (334.25G, 340.80G, and 292.35G MACs) compared with typical CNN/GAN-based methods. Overall, many state-of-the-art face restoration methods still require relatively high computational cost in terms of MACs and inference time, as shown in Table~\ref{table:speed}. From the perspective of model paradigm, GAN-based methods (\eg GFP-GAN, GPEN, and HiFaceGAN) are typically feed-forward networks and therefore can achieve relatively lower inference latency, while diffusion-based methods (\eg DifFace) rely on iterative denoising/sampling, which generally leads to longer inference time despite competitive restoration quality.

To better guide practical deployment, lightweight strategies can be considered from both architecture and optimization perspectives. On the architecture side, it is promising to use efficient backbones, reduce feature resolution in expensive blocks, and replace standard attention with efficient attention variants to lower memory and MACs. On the optimization side, pruning, quantization, and knowledge distillation can reduce model size and latency with limited quality drop. For diffusion-based methods, step reduction, fast samplers, and distilled sampling are practical directions to accelerate inference. In the future, we expect more work on developing lightweight face restoration models for edge devices.

\section{Future Directions}\label{sec:future_directions}
Despite great breakthroughs in face restoration technology, there still exist many challenges and unsolved problems. In this section, we discuss the limitations of existing methods and introduce new trends for future work.


\textbf{Network Design}. As discussed in the performance evaluation, the network structure can significantly influence the restoration performance. For example, recent Transformer-based methods usually have better performance due to the strong ability of Transformer architecture. GAN-based methods can generate visually pleasing face images with better non-reference metric values. Thus, when designing the network, it is worthwhile to learn from different structures, including CNN, GAN, ViT, and diffusion models. In particular, recent diffusion models provide a strong generative prior and have shown impressive capability in restoring realistic facial details. However, diffusion-based face restoration still faces several practical challenges, such as slow sampling speed, controllability of facial attributes, and identity consistency. Thus, future work could explore (i) more efficient diffusion sampling strategies (\eg step reduction and distillation) to accelerate inference, (ii) stronger conditional mechanisms (\eg conditioning on identity embeddings, landmarks, or 3D priors) to improve faithfulness, and (iii) controllable restoration strategies to balance realism and fidelity under different degradation levels. In addition, it is promising to investigate hybrid architectures that combine the efficiency of CNN/Transformer backbones with the strong prior of generative models to achieve both high quality and fast inference. Besides purely data-driven designs, another physically informed framework for image restoration is deep unfolding, which unrolls an iterative optimization procedure into a learnable multi-stage network.  Recent works such as DGUNet~\cite{mou2022deep} and UnfoldLDM~\cite{he2025unfoldldm} demonstrate its effectiveness for blind restoration and its synergy with diffusion priors. These advances suggest deep unfolding is promising for blind/real-world face restoration, and future work may incorporate face-specific priors (\eg identity/landmarks/geometry) into unfolding stages to better preserve faithfulness and identity consistency.

\textbf{Integration of Facial Priors and Networks}. As a domain-specific image restoration task, the facial features can be used in the face restoration task. When designing models, many methods aim at exploiting facial priors to recover realistic face details. Although some methods attempt to introduce geometry prior, facial component, generative prior, or 3D prior into face restoration, how to integrate the prior information into networks is still a promising direction for this task. Meanwhile, with the development of large-scale pre-training, advanced pre-trained models can provide powerful face-related priors learned from massive data. For instance, priors from diffusion models or self-supervised models (\eg DINO, MAE) can be incorporated via latent-space constraints, feature-space guidance, or plug-in/adaptor modules to improve generalization under real-world degradations.

\textbf{Loss Function and Evaluation Metrics}. For the face restoration task,  different loss functions have been adopted in the literature. The widely-used loss functions are \text{L1} loss, \text{L2} loss, perceptual loss, adversarial loss, and face-specific loss. Instead of using a single loss function, existing methods usually combine multiple loss functions with corresponding weights to train models. However, it is still not clear how to develop the right loss function for guiding the model training. Thus, in the future, more works are expected to seek more accurate loss functions (\eg general or task-driven loss functions) to promote the development of face restoration. In addition, loss functions can directly influence the evaluation results of models. As shown in Table~\ref{table:128_1} and \ref{tab:merged}, the pixel-wise \text{L1} loss and \text{L2} Loss tend to obtain better results in terms of PSNR, SSIM, and MS-SSIM. The perceptual loss and adversarial loss tend to generate more visual-pleasing results (\ie producing high LPIPS, FID, and NIQE values). Thus, how to develop metrics that can consider both human and machine aspects for model evaluation is also an important direction in the future.

\textbf{Computational Cost}. Existing face restoration methods aim at improving the restoration performance by significantly increasing the depth or width of the network, ignoring the computational cost of models. The heavy computational cost prevents these methods from being used in resource-limited environments, such as mobile or embedded devices. For example, as shown in Table~\ref{table:speed}, the state-of-the-art method RestoreFormer~\cite{wang2022restoreformer}
has $72.37$M parameters and $340.80$G MACs. The diffusion-based method DifFace suffers from a longer inference speed. It is very difficult to
deploy them in real-world applications. Therefore, developing models with a lighter computational cost is an important future direction.

\textbf{Standard Benchmark Datasets}. Unlike other low-level visual tasks such as image deblurring, image denoising, and image dehazing, there are few standard evaluation benchmarks for face restoration~\cite{zhang2022blind}. For example, most face restoration methods~\cite{li2020blind,wang2021towards,wang2022restoreformer} conduct experiments on private datasets (synthesizing the training set from FFHQ). 
Researchers may tend to use data that is biased to their proposed methods. On the other hand, to make a fair comparison, subsequent works need to take a lot of time to synthesize private data sets and retrain other comparison methods. In addition, the scale of the recent widely-used dataset is usually small, which is not suitable for deep learning methods. Thus, developing standard benchmark datasets is a direction for the face restoration task. In the future, we expect more standard and high-quality benchmark datasets to be built by researchers in the community.


\textbf{Video Face Restoration.}
With the popularization of mobile phones and cameras, the video face restoration task has become more and more important. However, existing works mainly focus on image-level face restoration, and video-related face restoration remains relatively under-explored, with only a limited number of representative studies and benchmarks reported in recent years~\cite{xu2024beyond,chen2024towards,wang2025svfr}. On the other hand, other low-level visual tasks such as video deblurring, video super-resolution, and video denoising have developed rapidly in recent years~\cite{wang2019edvr,chan2021basicvsr,chan2022basicvsr++,tassano2020fastdvdnet}. Therefore, video face restoration is a potential direction for the community. The task of video face restoration can be considered from the following two aspects. First, for the benchmark dataset, we could consider building high-quality video datasets for this task, which can quickly facilitate algorithm design and evaluation and benefit the community of face restoration, for example recent efforts on real-world video face restoration benchmarks~\cite{chen2024towards}. Second, for video restoration methods, we should develop video face restoration by fully considering the spatial and temporal information among successive frames to improve temporal consistency and reduce flickering artifacts, as explored in recent video face restoration models~\cite{xu2024beyond,wang2025svfr}.


\textbf{Real-world Face Restoration and Application}. Existing methods rely heavily on synthetic data to train networks. However, models trained on synthetic degradations do not necessarily generalize well to real-world scenarios. As shown in Table~\ref{tab:merged} and Fig.~\ref{fig:real_results}, most face restoration methods produce poor results on real-world face images due to the inherent domain gap between synthetic and real degradations. To alleviate this issue, some solutions have been proposed, such as unsupervised techniques or learning-based real degradation modeling. Nevertheless, many of these methods still depend on restrictive assumptions (\eg similar degradation patterns across images), making real-world deployment a continuing challenge. To further bridge the synthetic-to-real domain gap, advanced learning paradigms may empower real-world face restoration, including self-supervised/unsupervised learning, degradation modeling with richer and more diverse real degradations, and domain adaptation techniques that reduce reliance on paired training data. Another practical direction is to explicitly estimate degradation factors (\eg blur/noise/compression characteristics) and use them as conditions for restoration, which may improve robustness when degradations are unknown or mixed in the wild. In addition, some methods~\cite{shen2020exploiting,jiang2020dual} have shown that face restoration can improve the performance of subsequent tasks such as face verification and face recognition. However, how to couple face restoration with these tasks within a unified framework remains an important future research direction.


\textbf{Other Related Tasks}. In addition to the above-discussed face restoration tasks, there are many tasks related to face restoration, including face retouching~\cite{shafaei2021autoretouch,lei2022abpn}, photo-sketch synthesis~\cite{zhu2021sketch,yu2022efficient}, face-to-face translation~\cite{kr2019towards,yang2022s2fgan}, color enhancement~\cite {zhou2022towards,ji2022colorformer}, and old photo restoration~\cite{wan2020bringing,wan2022old}. For example, face retouching entails digitally enhancing facial features and appearance in photographs to create a more attractive and polished look. The goal of face retouching is to attain smoother skin, enhance complexion, improve the appearance of the eyes and teeth, and make adjustments to hair, makeup, and facial proportions. Old photo restoration is the task of repairing old photos, where the degradation of old photos is rather diverse and complex (\eg noise, blur, and color fading). 
In addition, some tasks focus on facial style transfer, such as face-to-face translation and facial expression analysis, which are different from face restoration. Thus, applying the existing face restoration methods to these related tasks is also a promising direction, which can trigger more applications to land.

\textbf{Bias and Ethical Issues}. Deep face restoration models may inherit dataset bias from the training data, which can lead to inconsistent restoration quality across different demographic groups and attribute categories. Prior methods~\cite{wang2021towards,wang2022restoreformer,zhou2022towards} has shown that generative models can amplify demographic imbalance when training sets are skewed~\cite{leyva2024demographic}. Similar concerns have been widely discussed in face related technologies, where demographic bias can affect reliability and fairness~\cite{yucer2024racial}. In addition, face restoration may introduce non authentic details during hallucination, which can be problematic when restored results are used in sensitive applications such as surveillance and forensics~\cite{li2025survey}. There are also privacy and consent concerns because many large-scale face datasets are collected from the internet and may not reflect informed consent in all cases~\cite{wang2024beyond}. Thus, future work should consider bias aware evaluation across groups and attributes, transparent reporting of dataset limitations, and deployment safeguards that avoid treating restored faces as ground truth evidence in high risk scenarios.

\section{Conclusion}\label{sec:conclusion}
In this work, we have systematically surveyed face restoration methods using deep learning. We discuss different degradation models, the characteristics of face images, the challenges of face restoration, and the core ideas in existing state-of-the-art methods, including geometric prior based methods, reference prior based methods, generative prior based methods, and non-prior based methods. After comprehensively reviewing face restoration methods, we discuss advanced techniques in face recovery methods from aspects of network architecture, basic block, loss function, and benchmark dataset. We also evaluate the representative methods on synthetic and real-world datasets. Finally, we discuss the future directions, including network design, metrics, benchmark datasets, applications, \etc  

\section*{Acknowledgement}
This work is funded in part by the National Natural Science Foundation of China (Grant No. 62372480, 62372223, and U24A20330), GuangDong Basic and Applied Basic Research Foundation (2025A1515011361), Shenzhen Science and Technology Program (JCYJ20240813110459017),  the Key R\&D Program of Xinjiang Uygur Autonomous Region (Grant No. 2025B03043-1), and Nanjing University-China Mobile Communications Group Co., Ltd. Joint Institute.

\bibliographystyle{ACM-Reference-Format}
\bibliography{myreferences}

\end{document}